\documentclass[10pt,twocolumn,letterpaper]{article}

\usepackage[pagenumbers]{cvpr} 

\usepackage{booktabs,array}
\usepackage{colortbl}
\usepackage{tabularx}
\usepackage{multirow}
\usepackage{wrapfig}
\usepackage{graphicx}
\usepackage{amsmath,amssymb,amsfonts}
\usepackage{algorithmic}
\usepackage{graphicx}
\usepackage{textcomp}
\usepackage[table,x11names,dvipsnames,table]{xcolor}
\usepackage{color}
\usepackage{xspace}
\usepackage[skip=8pt]{caption}
\definecolor{MyDarkGray}{rgb}{0.5,0.5,0.5}

\usepackage{bbding}
\newcolumntype{C}[1]{>{\centering\arraybackslash}p{#1}}
\newcommand{\ru}{\rule{0mm}{3mm}}

\definecolor{cvprblue}{rgb}{0.21,0.49,0.74}
\usepackage[pagebackref,breaklinks,colorlinks,citecolor=cvprblue]{hyperref}

\title{Raising the Bar of AI-generated Image Detection with CLIP}

\author{Davide Cozzolino\textsuperscript{1} \ \ \ 
Giovanni Poggi\textsuperscript{1} \ \ \ 
Riccardo Corvi\textsuperscript{1} \ \ \ 
Matthias Nie\ss ner\textsuperscript{2} \ \ \ 
Luisa Verdoliva\textsuperscript{1,2} \\[2mm]
{\textsuperscript{1}University Federico II of Naples \ \ \ \ \ \textsuperscript{2}Technical University of Munich}}

\begin{document}
\maketitle

\begin{abstract}
The aim of this work is to explore the potential of pre-trained vision-language models (VLMs) for universal detection of AI-generated images. We develop a lightweight detection strategy based on CLIP features and study its performance in a wide variety of challenging scenarios. We find that, contrary to previous beliefs, it is neither necessary nor convenient to use a large domain-specific dataset for training. On the contrary, by using only a handful of example images from a single generative model, a CLIP-based detector exhibits surprising generalization ability and high robustness across different architectures, including recent commercial tools such as Dalle-3, Midjourney v5, and Firefly.
We match the state-of-the-art (SoTA) on in-distribution data and significantly improve upon it in terms of generalization to out-of-distribution data (+6\% AUC) and robustness to impaired/laundered data (+13\%).
Our project is available at
{\small \url{https://grip-unina.github.io/ClipBased-SyntheticImageDetection/}}
\end{abstract}

\section{Introduction}
\label{sec:intro}

Synthetic images have by now left research laboratories and are flooding the real world.
The latest versions of popular image editing tools, such as Adobe Photoshop and Microsoft Paint, come with easy-to-use AI-powered generative tools that allow even novice users to edit and generate visual content at will.
Thanks to diffusion-based generative models, it is not only the quality of the generated images that is greatly improved, but also the flexibility in using these tools.
By issuing a few text commands you can easily obtain the desired image.
While this represents a great opportunity for visual arts applications, it is also the paradise of disinformation professionals who can design their attacks with unprecedented power and flexibility \cite{Cardenuto2023age, Barrett2023identifying, Epstein2023art}.
And, of course, a nightmare for all those trying to combat the spread of fake news and orchestrated disinformation campaigns: journalists, fact-checkers, law enforcement, governments. Therefore, there is a high demand for automatic tools that help establish the authenticity of a media asset.

\begin{figure}[t!]
    \centering
   \includegraphics[width=\linewidth]{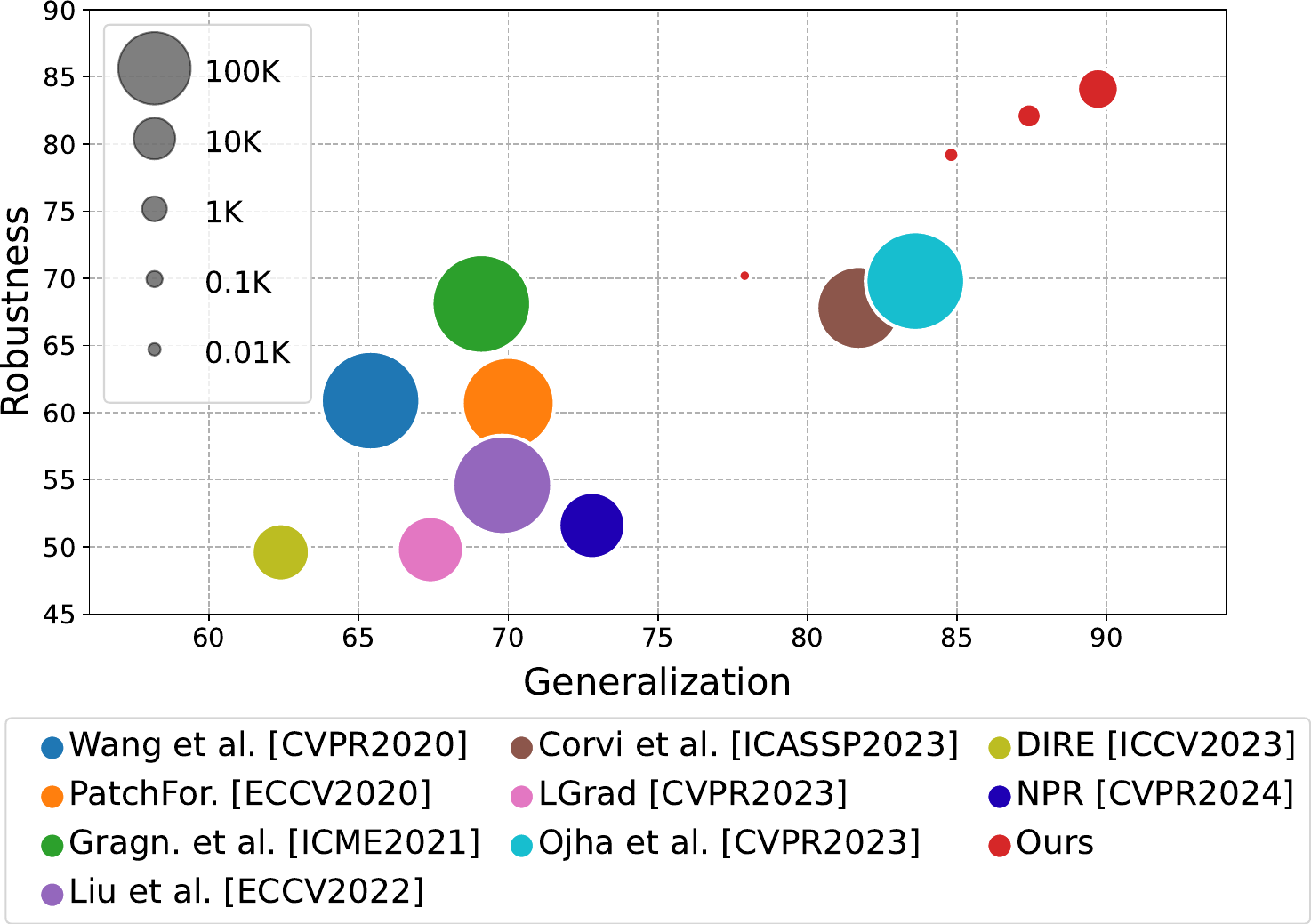}
    \caption{Area Under ROC Curve (AUC \%) on unseen synthetic generators ($x$-axis) and on post-processed data ($y$-axis). The first number measures the generalization ability of the detector, the second measures its robustness to possible impairments. Circle area is proportional to training set size. Performance is measured over 18 different synthetic models. Our CLIP-based detector largely outperforms all SoTA methods with very limited training data.}
    \label{fig:teaser}
\end{figure}

It is well known that each synthesis model leaves its own peculiar traces in all generated images, subtle traces that give rise to so called {\em artificial fingerprints}, that can be exploited for forensic analyses \cite{marra2019DoGAN, yu2019attributing}.
This phenomenon was first observed for methods based on generative adversarial networks (GAN) \cite{zhang2019detecting, durall2020watch} but holds, with obvious differences, for all generation approaches, including the more recent diffusion models (DM) \cite{corvi2023intriguing}.
One of the main challenges for current forensic detectors is the ability to generalize to new and unseen generative methods. Indeed, the in-distribution (ID) scenario, with perfectly aligned training and test sets, is rarely met in practice.
Test images are often generated by new approaches, unseen architectures, or even known architectures re-trained under different conditions, all representing different flavors of the out-of-distribution (OOD) scenario. Additionally, most images are downloaded from social networks, where they undergo transformations such as compression and resizing (sometimes multiple times), processes that tend to wash out the tiny traces so precious for detection.
In \cite{wang2020cnn}, it is shown that training diversity and intense augmentation are crucial for generalization. A ResNet-50 detector is trained on a single but highly diverse dataset of about 360k ProGAN \cite{karras2018progressive} images.
Results on OOD images generated by other GAN-based generators turned out to be surprisingly good. However, the authors themselves attribute this performance to the structural similarities between GAN-based generators.
Indeed, results are not equally good on images generated by recent diffusion-based methods, which present somewhat different generation artifacts \cite{corvi2022detection}.

Generalizing to OOD data is a major issue in deep learning and has been the object of intense research in recent years. In this context, the advent of large pre-trained vision-language models has brought about a number of new solutions and exciting results.
These models have been shown to be excellent zero-shot and few-shot learners in many diverse applications, such as image classification \cite{Radford2021learning, zhang2022tip-adapter}, detection \cite{esmaeilpour2022zero, ming2022delving} and segmentation \cite{xu2022simple, zhou2023zegclip}.
Recently, there have been attempts to exploit the power of VLMs to detect synthetic images \cite{sha2022fake, amoroso2023parents, ojha2023towards}.
For example, \cite{ojha2023towards} considers the same dataset, augmentation strategy and experimental protocol as in \cite{wang2020cnn} but uses a pre-trained VLM, the Contrastive Language-Image Pre-Training (CLIP) \cite{Radford2021learning}, as a feature extractor
rather than training a ResNet-50.
Only the classifier is then learned on the task-specific dataset.
Compared to \cite{wang2020cnn} the performance improves significantly, especially on images generated by diffusion models, showing excellent generalization ability.

In this work, we explore in depth the potential of CLIP for image forensics and conduct an extensive experimental analysis in challenging real-world scenarios involving a large number of generative models.
We find that CLIP-based methods show impressive generalization ability and very good robustness, resulting in large performance improvements compared to the state of the art.
To achieve such results, we avoid any intensive training on domain-specific data, which could introduce unwanted biases and undermine the descriptive power of CLIP features.
Instead, we rely on a small number of paired real/fake images with the same textual description and use their CLIP features to model the decision space. Synthetic images come from a single generator, but results are equally good across all different data sources.
The top performance is achieved with just 1,000 to 10,000 paired images. Moreover, only minor performance decays are observed when this number reduces to 100 or even 10 (see Fig. \ref{fig:teaser}). In summary, the key contributions of this work are as follows:
\begin{itemize}
\item We show that CLIP features achieve excellent generalization: by exploiting only a handful of examples, not even belonging to the generator under test, the performances are comparable to those of intensively trained solutions.
\item We carry out a large set of experiments on diverse synthetic generators and in very challenging conditions achieving the best performance on average. Our experiments make clear that the features extracted are partially orthogonal to the low-level features used by previous methods.
\end{itemize}

\section{Related work}
\label{sec:related}

There is an extensive literature on synthetic image detection, primarily focusing on images created by generative adversarial networks (GANs) and, more recently, by diffusion models (DMs).
Some methods look for visible errors, such as asymmetries in faces, incorrect perspectives, or unusual shadows \cite{matern2019exploiting, farid2022lighting, farid2022perspective}.
However, with the rapid progress in image synthesis technology, these types of problems are solved quickly and appear less frequently in modern generation methods.
Therefore, we will focus on methods that exploit ``invisible'' forensic traces, working either in the spatial or frequency domain.
Subsequently, we will provide a brief overview of the most recent approaches that exploit multimodal features.

\vspace{2mm} \noindent
{\bf Spatial domain methods.}
Despite their high visual quality, synthetic images carry with them distinctive traces of the generation process that enable detection and even pattern identification \cite{marra2019DoGAN, yu2019attributing}.
Each generation model, in fact, inserts a sort of digital fingerprint into all the images it creates, which depends on its architectural and training details.
This fingerprint can be easily estimated. Given a few hundred images generated by the target model, it is sufficient to average their noise residuals, extracted using simple denoisers \cite{marra2019DoGAN} or more sophisticated methods based on deep learning \cite{sinitsa2023deep, liu2022detecting}.
Detectors based on digital fingerprints can be regarded as ``few-shot'', as they can deal with new models based on just a few example images. Other few-shot solutions have been proposed lately in this field \cite{cozzolino2018forensictransfer, marra2019incremental, du2020towards, jeon2020tgd}.
However, they all require information on the target models, even if limited to a few images, so they fit a strictly in-distribution scenario.
In contrast, our work aims to generalize to new and previously unseen models, and therefore fits into an out-of-distribution scenario.

Good generalization is a key requirement of synthetic image detectors.
Towards this end, it is important to increase diversity in training, which can be pursued by suitable augmentation, by including a large number of different categories \cite{wang2020cnn}, or even by model ensembling \cite{mandelli2022detecting}.
Other studies suggest working on local patches \cite{chai2020what} or combine global spatial information with local features \cite{ju2022fusing}.
In \cite{gragnaniello2021GAN}, it is shown that to preserve the subtle high-frequency forensic traces, one should avoid any downsampling in the first layers of the network.
With the same aim, \cite{tan2023learning} proposes to work on noise residuals rather than the original images, extracting the gradients with a pre-trained CNN.
In \cite{chandrasegaran2022discovering}, a systematic study is carried out on transferable forensic features that allows easier generalization.

The above investigations and proposals, however, only consider GAN-based generators.
Some very recent works extend the analyses to latest generation approaches. \cite{corvi2022detection} shows that detectors designed for GAN images have a hard time generalizing to DM images, especially in the presence of common post-processing steps such as compression and resizing.
Furthermore, they are unable to select an adequate decision threshold without the help of some calibration data from the model under test.
On the other hand, \cite{epstein2023online} shows that it is possible to achieve decent performance by continuously re-training the detector on images from new generators, as long as the latter share some architectural components with the known ones.
In \cite{wang2023dire}, inspired by previous work on GAN images \cite{Albright2019source}, a detector specifically tailored to DM images is proposed. Images are projected in a latent space and reconstructed with known models to study their features. Although promising, the approach refers to specific architectures, hence an ID scenario, and has been tested on a limited set of generative diffusion models.
In this work instead, we do not make any assumptions and include a large variety of both GAN and DM generators.

\vspace{2mm} \noindent
{\bf Frequency domain methods.}
GAN-image artifacts are more easily spotted in the frequency domain and are clearly visible in the artificial fingerprint spectra \cite{durall2020watch, frank2020leveraging, dzanic2020Fourier}.
In fact, they are caused by the up-sampling operations used in the generator, which give rise to regular spatial patterns and strong peaks in the Fourier domain.
Interestingly, the awareness of such weakness has prompted the design of new architectures, like StyleGAN3 \cite{karras2021alias}, which explicitly avoid aliasing and reduce the above-mentioned peaks.
In addition to such obvious artifacts, the spectral content of GAN images and real images is also known to differ significantly in the medium and high frequencies.
Likewise, clear differences in the radial and angular spectral distributions of the DM and real images were observed \cite{corvi2023intriguing, yang2022diffusion}.
Furthermore, it is worth observing that DM images may also present spectral peaks (see Fig.~\ref{tig:beyond}).
Of course, frequency domain traces can be used to train simple detectors for ID images.
Even more interestingly, some authors \cite{zhang2019detecting, jeong2022fingerprintnet} replicate and modify the synthesis process of known generators by introducing a series of small architectural changes to learn to handle even OOD test images.
A major problem with spectral traces is their low resilience to laundering operations, both unintentional (e.g. image resizing) and intentional (e.g., counterforensic methods \cite{dong2022think,chandrasegaran2021closer, cozzolino2021spoc}).

\vspace{2mm} \noindent
{\bf Methods based on multimodal features.}
The introduction of large language models has sparked intense work in the vision community attempting to leverage large models trained using both images and text \cite{zhou2022learning,zhang2022tip-adapter}.
In image forensics, however, traditionally oriented towards the analysis of low-level features, there has been very little work in this direction. Only a few papers \cite{sha2022fake, amoroso2023parents, ojha2023towards} have tried to leverage pre-trained visual models for AI-generated image detection.
All these studies rely on variants of CLIP.
However, only in \cite{ojha2023towards} is generalization to OOD data pursued, via nearest neighbor and linear probing in CLIP feature space.
To this end, a large dataset of fake and real images is used to train the classifier.
In this work we take a further step in this direction and show that superior performance can be achieved even using much less data.

\begin{figure}[t!]
	\centering
	\setlength{\tabcolsep}{0.2em}
	\begin{tabular}{ccc}
		\includegraphics[width=2.5cm]{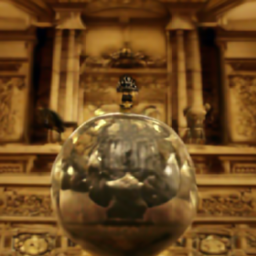} &
            \includegraphics[width=2.5cm]{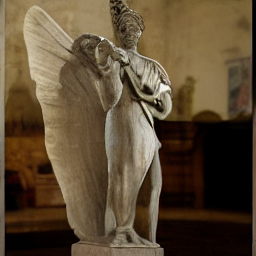} &
		\includegraphics[width=2.5cm]{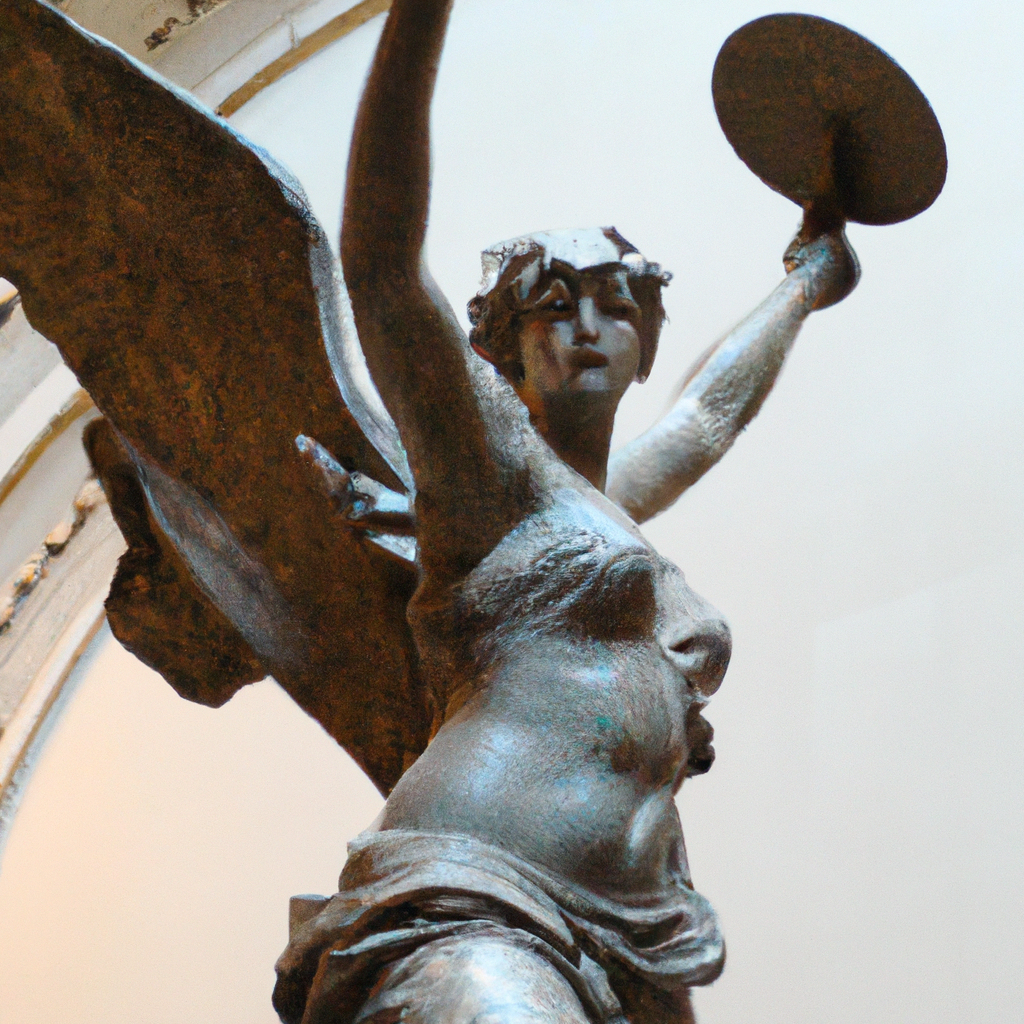}
  		\vspace{0.15cm} \\
		\includegraphics[width=2.5cm]{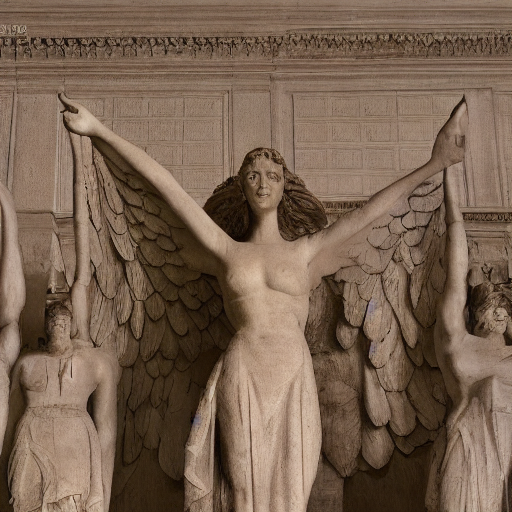} &
  		\includegraphics[width=2.5cm]{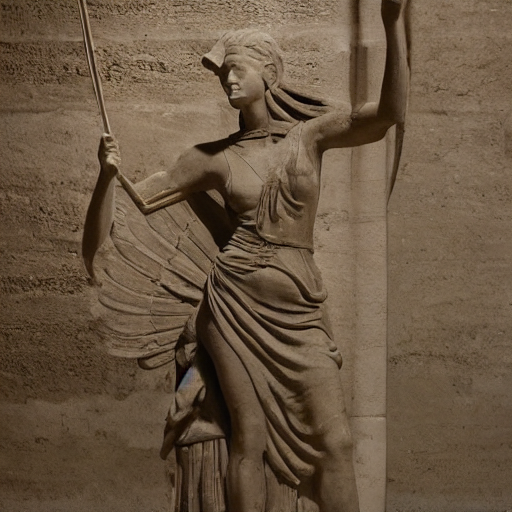} &
            \includegraphics[width=2.5cm]{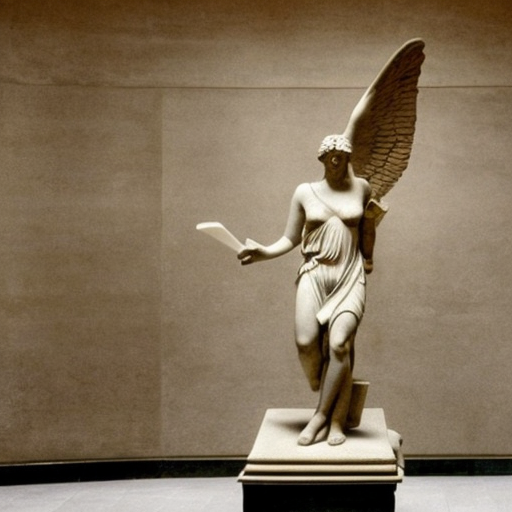}
		\vspace{0.15cm} \\
            \includegraphics[width=2.5cm]{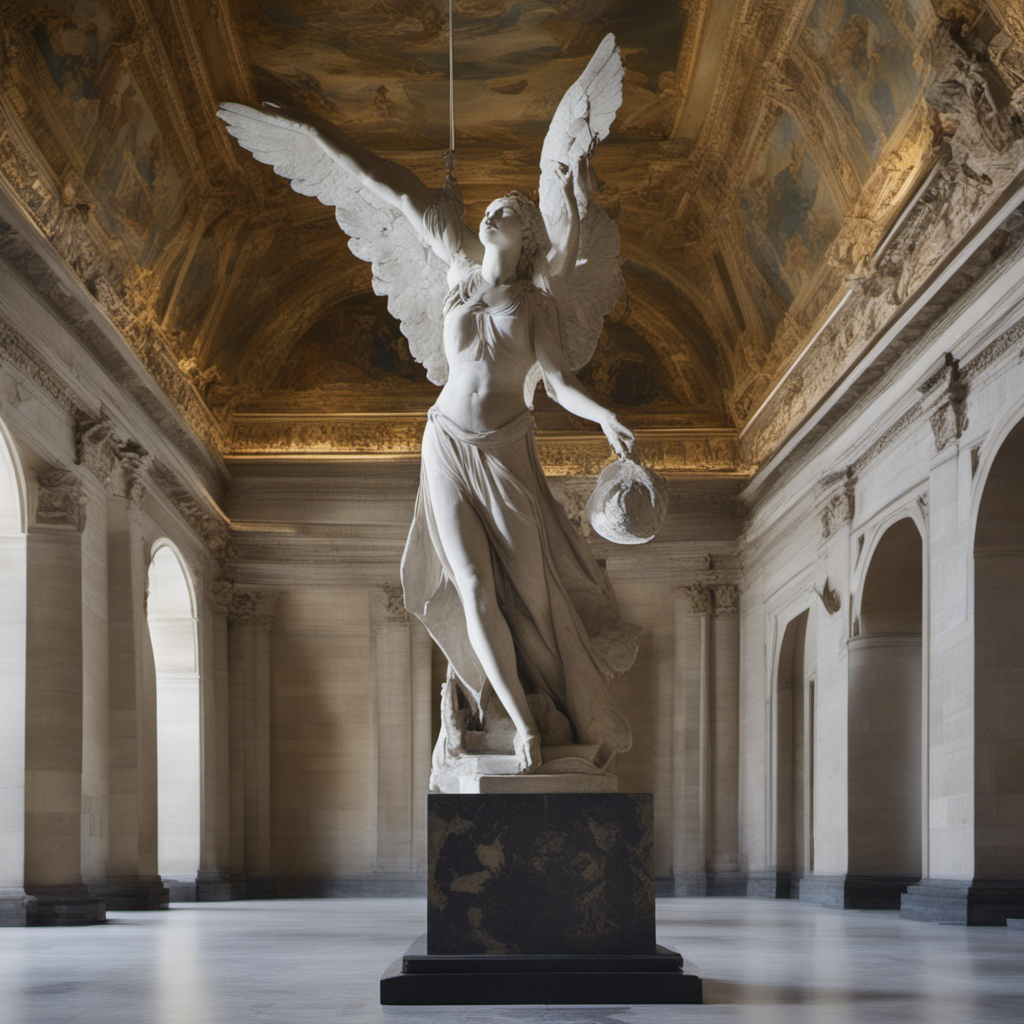} &
            \includegraphics[width=2.5cm]{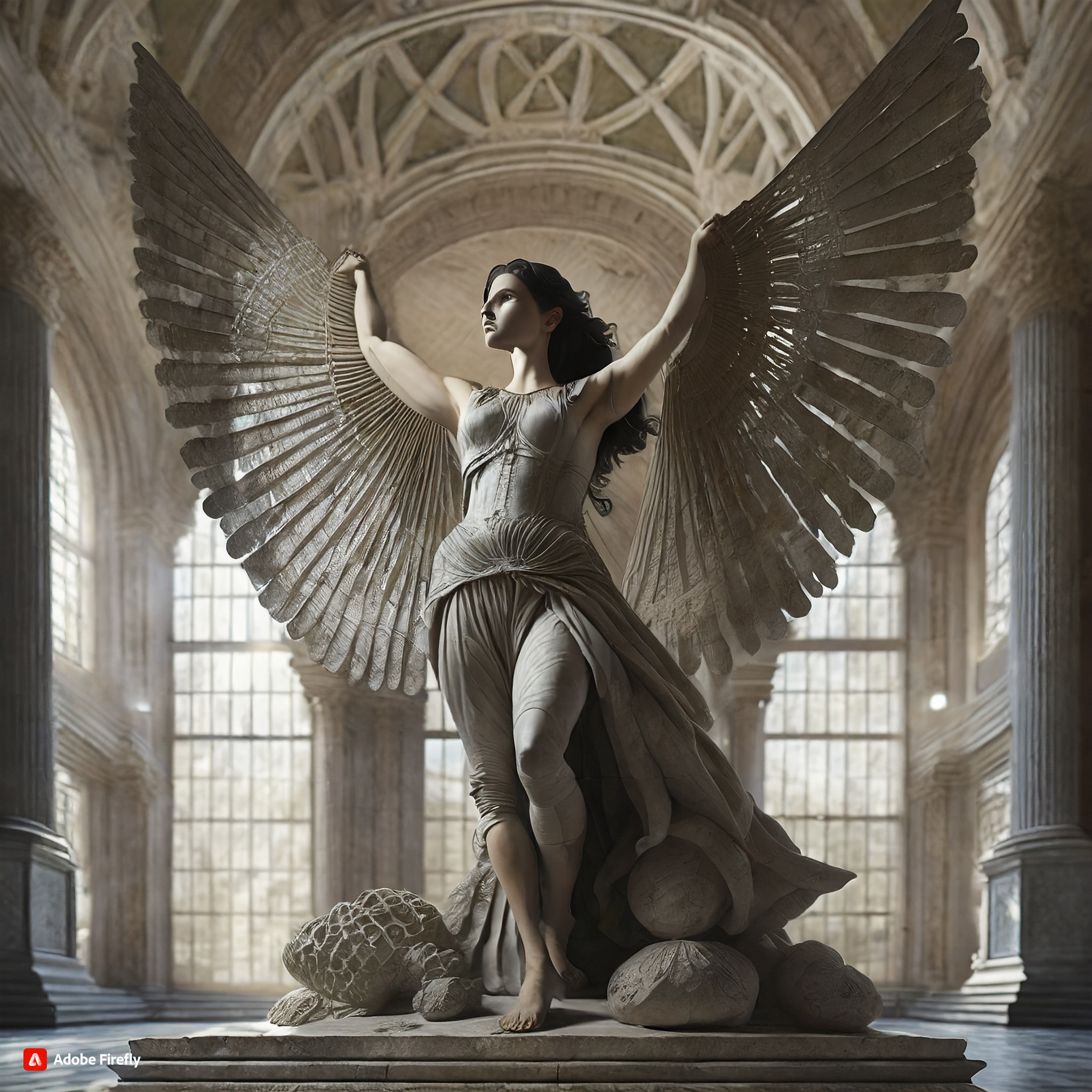} &
            \includegraphics[width=2.5cm]{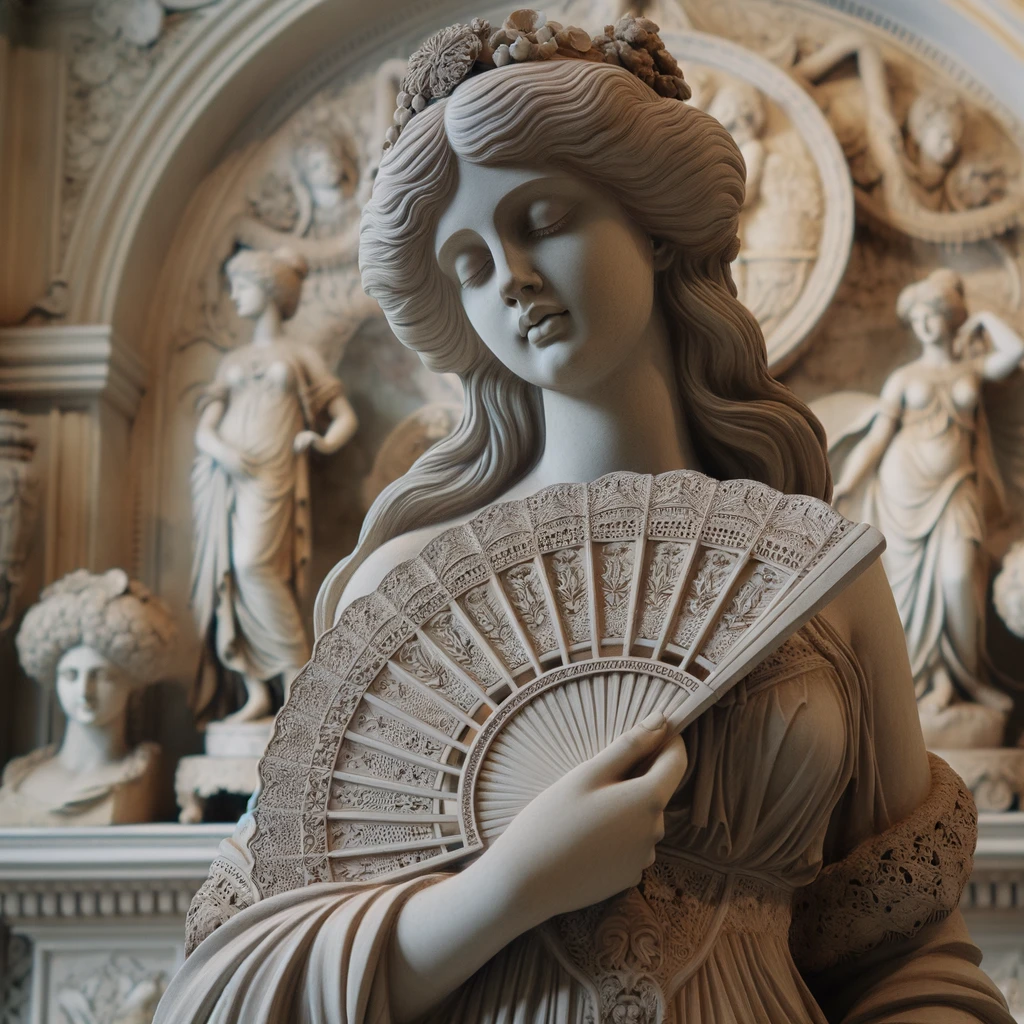}
	\end{tabular}
	\caption{Examples of synthetic images from generators used in our experiments.
    From left to right, Top: GLIDE \cite{nichol2021glide}, Latent Diffusion \cite{ramesh2022hierarchical}, DALL·E 2 \cite{ramesh2022hierarchical}. Middle: Stable Diffusion 1.3, Stable Diffusion 1.4, Stable Diffusion 2.1 \cite{stablediffusion2}. Bottom: Stable Diffusion XL \cite{podell2023sdxl}, Adobe Firefly \cite{firefly}, DALL·E 3 \cite{dalle3}.
    }
 \label{fig:examples}
\end{figure}

\section{Datasets and metrics}
\label{sec:datasets}

The main objective of this work is to study the generalization capacity of conventional and CLIP-based methods, i.e. to evaluate how detectors behave on data never seen before, of different origins and possibly impaired by various forms of post-processing.
Therefore, we consider a very large test dataset and take measures to mitigate any potential biases in the experimental validation process.

Tab.~\ref{tab:datasets} lists all generation models used during test.
They are grouped into three families:
GAN-based (ProGAN, StyleGAN2, StyleGAN3, StyleGAN-T, GigaGAN),
DM-based (Score-SDE, ADM, GLIDE, eDiff-I, Latent and Stable Diffusion, DiT, DeepFloyd-IF, Stable Diffusion XL)
and commercial tools (DALL·E 2, DALL·E 3, Midjourney v5 and Adobe Firefly), which include some recent text-based methods available online, whose architecture is not always disclosed.
The second column of the table specifies the generation modalities employed by these models: unconditional (u), conditional (c), and text-to-image (t).
The last column reports the resolution of the images in the dataset.
The central columns, indicate which datasets of real images were considered to carry out the detection experiments.
These have been carefully selected to avoid all foreseeable biases.
For example, when testing synthetic images generated by ProGAN, we use as real counterparts images from the LSUN dataset on which the ProGAN model was trained.

The StyleGAN-T~\cite{sauer2023stylegan}, eDiff-I~\cite{balaji2022ediffi}, and GigaGAN~\cite{kang2023scaling} images were provided by the authors of the respective papers.
Instead, we generated the images ourselves for Score-SDE~\cite{song2020score}, Stable Diff.~\cite{stablediffusion2}, DiT~\cite{peebles2022scalable}, and Deepfloyd-IF~\cite{deepfloydif} using the pre-trained models available online.
Additional generated images were sourced from two public datasets \cite{wang2020cnn, corvi2022detection}.
For the text-driven commercial tools, we used a recently proposed dataset \cite{bammey2023synthbuster}.
Pristine and synthetic images with similar semantic content are built by extracting textual descriptions from a real dataset (RAISE \cite{nguyen2015raise}) and using them as prompts for generating synthetic images.
A few examples of such images are shown in Fig.~\ref{fig:examples}.

Overall, we have a dataset with $32,000$ real and fake images for all upcoming tests.
To consider a more realistic scenario we simulate images shared on social networks and subject to post-processing operations: random cropping that can vary in a range from $\frac{5}{8}$ to the full size of the image, resizing to $200\times200$ pixels, and JPEG compression with a random quality factor between $65$ and $100$.

To measure performance, we consider three metrics. The area under the receiver operating curve (AUC) and average precision (AP) are both threshold-independent. The Accuracy instead, given by the number of correct predictions divided by the  number of tested images, depends on the decision threshold.
We always use a fixed threshold of 0.5, to simulate a realistic scenario in which no prior information on the data under test is available to carry out calibration.

\newcommand{\twr}[1]{\multirow{2}{*}{\shortstack[c]{#1}}}
\newcommand{\thr}[1]{\multirow{3}{*}{\shortstack[c]{#1}}}
\newcommand{\cm}[0]{\checkmark}

\newcommand{\rob}[1]{\rotatebox{90}{#1}}
\begin{table}[t!]
\centering
\resizebox{1.\linewidth}{!}{
\def\arraystretch{0.8}
\begin{tabular}{@{}l|c|c|c|c|c|c|c|l@{}}
\ru Generator                                     & \rob{modality} & \rob{LSUN \cite{yu2015lsun}} &
\rob{FFHQ \cite{karras2019stylegan}} &\rob{ImageNet \cite{deng2009imagenet}}& \rob{COCO \cite{lin2014microsoft}} & \rob{LAION \cite{schuhmann2021laion}} & \rob{RAISE \cite{nguyen2015raise}} & Resolution                 \\ \midrule
\ru ProGAN          \cite{karras2018progressive}  &       u        & \cm        &            &              &            &             &             & $256^{2}$                  \\
\ru StyleGAN2       \cite{karras2020stylegan2}    &       u        & \cm        & \cm        &              &            &             &             & $256^{2}$-$1024^{2}$       \\
\ru StyleGAN3       \cite{karras2021alias}        &       u        &            & \cm        &              &            &             &             & $1024^{2}$       \\
\ru StyleGAN-T      \cite{sauer2023stylegan}      &       t        &            &            &              & \cm        &             &             & $512^{2}$                  \\
\ru GigaGAN         \cite{kang2023scaling}        &       c,t      &            &            & \cm          &            & \cm         &             & $256^{2}$,$512^{2}$        \\ \midrule
\ru Score-SDE       \cite{song2020score}          &       u        &            & \cm        &              &            &             &             & $256^{2}$                  \\
\ru ADM             \cite{dhariwal2021diffusion}  &       u,c      & \cm        &            & \cm          &            &             &             & $256^{2}$                  \\
\ru GLIDE           \cite{nichol2021glide}        &       t        &            &            &              & \cm        &             & \cm         & $256^{2}$                  \\
\ru eDiff-I         \cite{balaji2022ediffi}       &       t        &            &            &              & \cm        &             &             & $256^{2}$,$1024^{2}$       \\
\ru Latent Diff.    \cite{rombach2022high}        &       u,c,t    & \cm        & \cm        & \cm          & \cm        &             &             & $256^{2}$                  \\
\ru Stable Diff.    \cite{stablediffusion2}       &       t        &            &            &              &            & \cm         & \cm         & $256^{2}$-$768^{2}$        \\
\ru DiT             \cite{peebles2022scalable}    &       c        &            &            & \cm          &            &             &             & $256^{2}$,$512^{2}$        \\
\ru DeepFloyd-IF    \cite{deepfloydif}            &       t        &            &            &              & \cm        &             &             & $1024^{2}$                 \\
\ru Stable Diff. XL \cite{podell2023sdxl}         &       t        &            &            &              &            &             & \cm         & $1024^{2}$                 \\ \midrule
\ru DALL·E 2        \cite{ramesh2022hierarchical} &       t        &            &            &              &            &             & \cm         & $1024^{2}$                 \\
\ru DALL·E 3        \cite{dalle3}                 &       t        &            &            &              &            &             & \cm         & $1024^{2}$                 \\
\ru Midjourney V5   \cite{midjourney}             &       t        &            &            &              &            &             & \cm         & $1024^{2}$-$1100^{2}$      \\
\ru Adobe Firefly   \cite{firefly}                &       t        &            &            &              &            &             & \cm         & $2032^{2}$-$2048^{2}$      \\
\bottomrule
\end{tabular}
}
\caption{Image generators used in our experiments: GAN-based, DM-based, and commercial tools.
For unconditional (u) and conditional (c) models, the real images used in test come from the same datasets used to train the generator, while for text-to-image (t) models synthetic images have been generated using the prompt extracted from the real counterpart.}
\label{tab:datasets}
\end{table}

\section{CLIP for synthetic image detection}
\label{sec:method}

\renewcommand{\r}{{\bf r}}
\newcommand{\f}{{\bf f}}
\newcommand{\CLIP}{{\rm CLIP}}

We propose a simple procedure to distinguish real images from synthetic images based on features extracted from the image encoder of CLIP ViT L/14.
The design of the detector consists of the following four steps:
\begin{enumerate}
\item   collect $N$ real images $\{R_1,\ldots,R_N\}$ with the corresponding captions $\{t_1,\ldots,t_N\}$;
\item   use the captions to feed a text-driven image generator, $G(\cdot)$, so as to obtain $N$ synthetic images $\{F_1,\ldots,F_N\}$ with $F_i=G(t_i)$. Now we have $N$ real / fake pairs that share the same textual description;
\item   feed CLIP with the $N$ real and $N$ fake images and collect the corresponding feature vectors $\{\r_1,\ldots,\r_N\}$ and $\{\f_1,\ldots,\f_N\}$ extracted at the output of the next-to-last layer, with $\r_i=\CLIP(R_i)$ and $\f_i=\CLIP(F_i)$;
\item   use these $N$+$N$ vectors to design a linear SVM classifier.
\end{enumerate}

If the real images have low-quality associated captions, or no caption at all, these can be generated by a dedicated tool such as BLIP \cite{li2022blip}.
Real and fake images are coupled to avoid possible semantic biases, which proves beneficial when the number of images is very small, {\it e.g.,} $N=10$.
We chose to extract features from the next-to-last layer rather than the last layer based on the results of preliminary experiments and, in the same way, we selected SVM among a set of candidate simple classifiers.
All these preliminary experiments and their results are described in the supplementary material.
In the following, we analyze the performance of the proposed detector as a function of the main design choices and experimental conditions.

\subsection{Influence of the reference set size}
\label{sec:training}

\begin{figure}[t!]
    \centering
    \includegraphics[width=\linewidth, page=1, trim=10 0 0 0, clip]{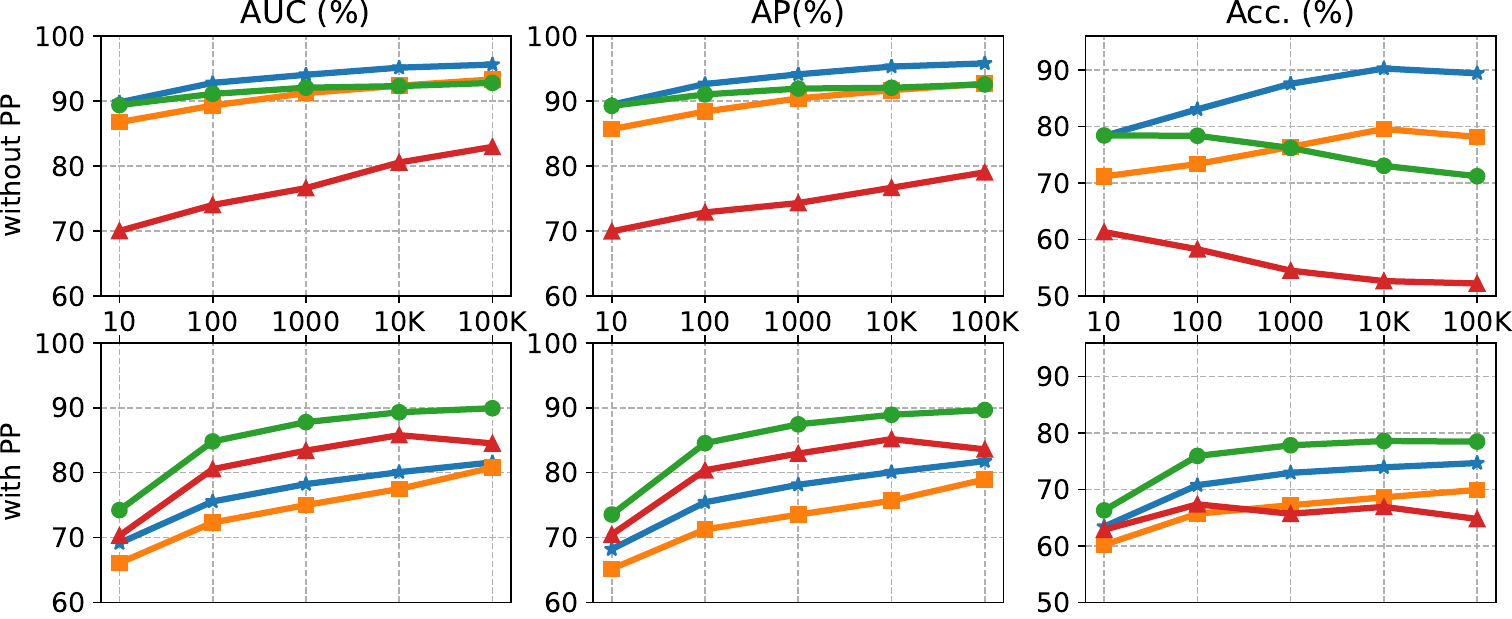} \\
    \includegraphics[width=\linewidth, page=2, trim=15 22 0 0, clip]{figures/results_size_training.pdf}
    \caption{Performance of the CLIP-based detector as a function of the number of real and synthetic images in the reference set.  We show AUC, AP and Accuracy on the original dataset (Top) and on post-processed images that simulate a realistic scenario (Bottom).}
    \label{fig:training-size}
\end{figure}

We start by investigating the role of the most important parameter of the proposed detector, the number of real / synthetic samples $N$ in the reference set, which in this experiment includes COCO real images and Latent Diffusion synthetic images.
Fig.~\ref{fig:training-size} shows results for $N$ ranging from 10 to 100k in log scale.
Results are in terms of AUC, AP and Accuracy.
When $N$ is less than 10k, we perform several runs with different sets of images and report the average metrics, {\it e.g.,} for $N$ equals 1k, we average on 10 runs.

Let’s focus first on the top row, which shows results on images that were neither recompressed nor resized.
In this simpler scenario, both AUC and AP values are extremely high, consistently above 85\% for all known families of synthetic images (GAN, Diffusion)
and significantly worse only for images generated by commercial tools.
It's particularly remarkable that very good results are achieved with as few as 10+10 reference images, proving that this lightweight solution is fully viable.
The performance improves with $N$, growing up to 5-10\% at the 10k plateau.
We note, in passing, that AUC and AP curves are very similar, so we drop the latter for brevity in further experiments.

In the rightmost figure, we see that, unlike AUC and AP, accuracy does not always improve as $N$ increases.
This is because AUC and AP are threshold-independent integral metrics.
They tell us how well a method with a perfectly calibrated threshold might work.
Instead, the precision depends critically on the selected threshold which, in the absence of any prior knowledge, is set at 0.5.
For Latent, 0.5 is a good threshold and accuracy improves as $N$ grows.
Instead, in general, the optimal threshold moves further away from 0.5 as $N$ increases and the precision decreases significantly.
So, there is a trade-off between optimality and robustness in our truly OOD scenario, where no calibration data exists.
On the other hand, if we had data from the target class, even just 10 images,
we could use them to design an {\it ad hoc} classifier (see supplementary material).

The bottom row of the figure shows results for images that have been compressed and/or resized.
As expected, performance degrades slightly in this scenario,
but continues to be very good, with AUC between 75\% and 90\% and accuracy between 65\% and 80\% at $N$=10k.
These are excellent results, well beyond the current state of the art, as we will show later.
In this case, using only 10+10 images does not seem advisable but the performance is almost optimal already at $N$=100.

\newcommand{\tworows}[1]{\multirow{2}{*}{\shortstack[c]{#1}}}
\begin{table}[t!]
{\small
\centering
\resizebox{1.\linewidth}{!}{
\begin{tabular}{l C{0.4cm}C{1.2cm}C{1.2cm}C{1.2cm}c}
\toprule
\tworows{Datasets}   &      augm. &    GAN & Diffusion & Commerc. & \tworows{AVG} \\
                     &            & family &    family &    tools &                \\ \midrule
\ru COCO $+$ Latent &     & 92.4 & 92.6 &  80.5 &  88.5  \\
\ru COCO $+$ Latent & \cm & 89.3 & 91.8 &  87.0 &  89.4  \\
\ru COCO $+$ ProGAN &     & 93.6 & 90.0 &  65.6 &  83.1  \\
\ru COCO $+$ ProGAN & \cm & 91.3 & 90.1 &  82.3 &  87.9  \\
\ru LSUN $+$ Latent &     & 88.7 & 82.7 &  65.6 &  79.0  \\
\ru LSUN $+$ Latent & \cm & 87.9 & 79.9 &  80.3 &  82.7  \\
\ru LSUN $+$ ProGAN &     & 94.1 & 79.2 &  44.7 &  72.7  \\
\ru LSUN $+$ ProGAN & \cm & 95.0 & 82.7 &  67.5 &  81.7  \\
\bottomrule
\end{tabular}
}
\vspace{-1.5mm}
\caption{AUC performance of the CLIP-based detector for various real/fake data, with and w/o augmentation (resizing/compression).}
\label{tab:content}
}
\end{table}

\subsection{Influence of the reference set content}
\label{sec:content}

We found that not only the quantity but also the quality of images in the reference set significantly impacts performance. This is intuitive with a very small reference set, like 10+10 images, where low quality or limited diversity could have catastrophic consequences. However, similar effects are observed even with a larger reference set.
Tab.~\ref{tab:content} shows the results in terms of AUC obtained using various combination of real and synthetic images in the reference set, that is, (Real, Synth) $\in \{ {\rm COCO, LSUN} \} \times \{ {\rm Latent, ProGAN}\}$, with and without augmentation.
In all cases, the reference set has size 10,000+10,000.
Results are clearly influenced by the specific combination.
Indeed, considerable degradation occurs when the LSUN dataset is used instead of COCO to draw the real samples and, to a lesser extent, when using ProGAN instead of Latent for the synthetic images.
We conjecture that both LSUN and ProGAN lack the diversity necessary to adequately describe the decision domain.
Furthermore, unlike COCO, the LSUN dataset exhibits several biases: the images represent only a few well-defined categories, have all the same size, and most of them are compressed with the same quality factor.
In Tab.~\ref{tab:content}, we see that data augmentation only marginally reduces the effects of the dataset used.

\begin{figure}
    \centering
    \includegraphics[width=\linewidth]{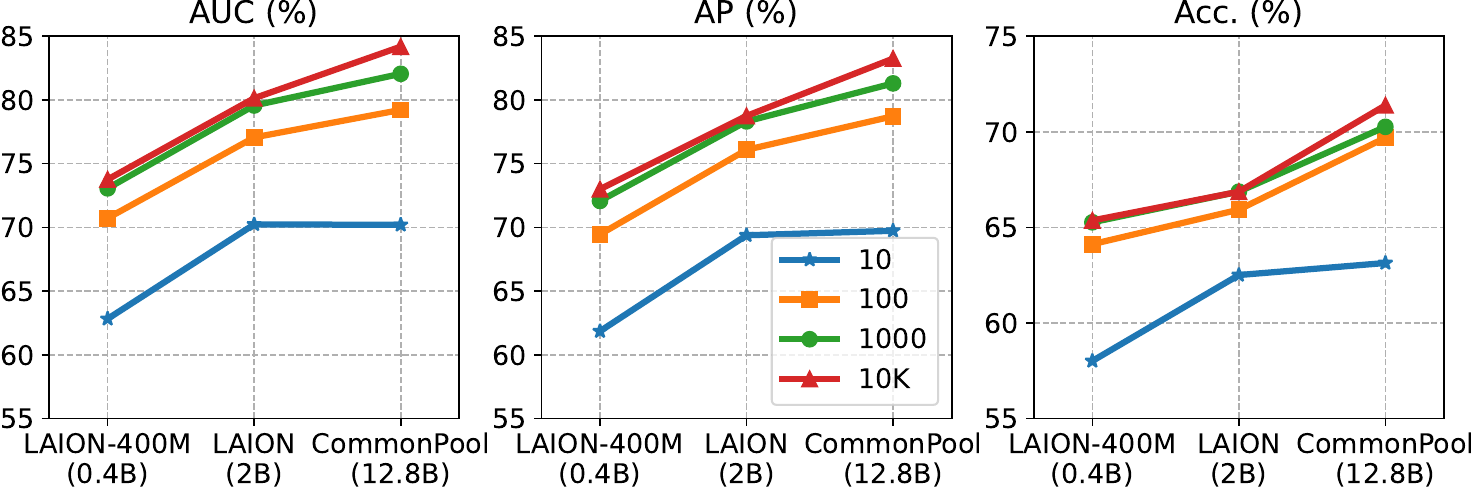}
    \caption{Performance of the CLIP-based detector as a function of the pre-training. We show AUC, AP and Accuracy on post-processed images for models pre-trained on LAION-400M (0.4B images), LAION (2B) and  CommonPool (12.8B).}
    \label{fig:pretraining-size}
\end{figure}

\subsection{Influence of pre-training}
\label{sec:pretrain}

We experimented also with various versions of CLIP ViT L/14 \cite{ilharco2021openclip}, trained on different datasets, observing significant changes in performance.
Fig.~\ref{fig:pretraining-size} shows, the AUC, AP and Accuracy averaged on all families of models as a function of the pre-training dataset.
Each curve refers to a different value of $N$ and we consider only the case of post processed images for brevity.
For both metrics and all values of $N$ there is a steady increase in performance as increasingly larger datasets are used for pre-training,
from LAION-400M (0.4B images) \cite{schuhmann2021laion}, to LAION (2B) \cite{schuhmann2022laionb}, to CommonPool (12.8B) \cite{gadre2023datacomp}.
Overall, there is a ten-point improvement from the smallest to the largest dataset.
This confirms the importance of pre-training a large VLM on the largest possible collection of different images.

\newcommand{\g}[1]{\textcolor{MyDarkGray}{#1}}
\renewcommand{\b}[1]{{\bf #1}}

\begin{table}[t!]
{\small
\centering
\resizebox{1.0\linewidth}{!}{
\begin{tabular}{l cccc} \toprule
     Method       & Real/Synth. Training &  Size (k)   & Aug. & Test Strategy   \\ \midrule
\ru \cite{wang2020cnn}         Wang et al.    &        LSUN / ProGAN &  360 / 360  & \checkmark & global pooling \\
\ru \cite{chai2020what}        PatchFor.      &  CelebA,FF / various &   84 / 272  &            & resizing       \\
\ru \cite{gragnaniello2021GAN} Grag. et al.   &        LSUN / ProGAN &  360 / 360  & \checkmark & global pooling \\
\ru \cite{mandelli2022detecting} Mand. et al. &    various / various &  232 / 386  & \checkmark & patch aggregation \\
\ru \cite{liu2022detecting}    Liu et al.     &        LSUN / ProGAN &  360 / 360  & \checkmark & global pooling \\
\ru \cite{corvi2022detection}  Corvi et al.   &   COCO,LSUN / Latent &  180 / 180  & \checkmark & global pooling \\
\ru \cite{tan2023learning}     LGrad          &        LSUN / ProGAN &   72 / 72   & \checkmark & resizing       \\
\ru \cite{ojha2023towards}     Ojha et al.    &        LSUN / ProGAN &  360 / 360  & \checkmark & cropping       \\
\ru \cite{wang2023dire}        DIRE-1         &       LSUN-Bed / ADM &   40 / 40   &            & resizing       \\
\ru \cite{wang2023dire}        DIRE-2         &    LSUN-Bed / St.GAN &   40 / 40   &            & resizing       \\
\ru \cite{tan2023rethinking}   NPR            &        LSUN / ProGAN &   72 / 72   &            & resizing       \\  \midrule
\ru Ours 1k       &        COCO / Latent &    1 / 1    &            & resizing       \\
\ru Ours 1k+      &        COCO / Latent &    1 / 1    & \checkmark & resizing       \\
\ru Ours 10k      &        COCO / Latent &   10 / 10   &            & resizing       \\
\ru Ours 10k+     &        COCO / Latent &   10 / 10   & \checkmark & resizing       \\
 \bottomrule
\end{tabular}
}
\vspace{-1.5mm}
\caption{{\bf List of methods.} For each method, we report the datasets of real and synthetic images used for training, their sizes, whether or not augmentation is used, and the testing strategy.}
\label{tab:sota_methods}
}
\end{table}

\newcolumntype{N}{>{\centering\arraybackslash\hsize=.09\hsize}X}
\begin{table*}[t!]
{\small
\centering
\resizebox{1.0\linewidth}{!}{
\begin{tabular}{l |C{7mm}C{7mm}C{7mm}C{7mm}C{7mm}|C{7mm}C{7mm}C{7mm}C{7mm}C{7mm}C{7mm}C{7mm}C{7mm}C{7mm}|C{7mm}C{7mm}C{7mm}C{7mm}|C{7mm}} \toprule
                  &\multicolumn{5}{c|}{GAN family}            & \multicolumn{9}{c|}{Diffusion family}
                  &\multicolumn{4}{c|}{Commercial tools} & AVG \\ \cmidrule(lr){2-6} \cmidrule(lr){7-15} \cmidrule(lr){16-19}
                  &      Pro &    Style &   Style  &   Style  &    Giga  &   Score  &\twr{ADM} &\twr{GLIDE}&  Latent  &  Stable  & DeepFl.  &\twr{Ediff-I}& \twr{DiT}&\twr{SDXL}& DALL· & DALL·  & \twr{Midj.} & Adobe   &    \\[-0mm]
     Method       &      GAN &     GAN2 &    GAN3  &    GANT  &     GAN  &     SDE  &          &          &   Diff.  &   Diff.  &      IF  &          &          &          &    E2    &     E3   &          &   Firef.       &   \\ \midrule
\ru Wang et al.   & \g{100.} &    96.5  &    98.5  &    98.9  &    66.6  &    32.9  &    64.3  &    48.5  &    59.2  &    41.5  &    78.0  &    64.9  &    58.6  &    54.3  &    64.8  &    10.9  &    40.2  &    84.8  &    64.6  \\
\ru PatchFor.     & \g{92.3} &    84.5  &    91.8  &    91.2  &    64.7  &    83.3  &    74.8  &    96.2  &    78.1  &    62.4  &    62.7  &    78.7  &    83.1  &    68.4  &    41.9  &    52.7  &    57.8  &    49.4  &    73.0  \\
\ru Grag. et al.  & \g{100.} & \b{99.8} &    97.5  &    98.8  &    82.8  &    92.1  &    74.7  &    62.8  &    91.9  &    52.5  &    69.9  &    69.6  &    65.3  &    58.0  &    58.3  &    ~2.4  &    43.1  &    63.5  &    71.3  \\
\ru Mand. et al.  &    96.2  & \g{93.8} & \b{100.} &    92.6  &    61.8  & \g{99.8} &    56.5  &    40.5  &    70.0  &    36.8  &    47.2  &    65.0  &    59.1  &    27.0  &    14.5  &    14.7  &    24.3  &    36.7  &    57.6  \\
\ru Liu et al.    & \g{100.} & \b{99.8} &    98.4  &    98.5  & \b{98.2} & \b{95.4} &    82.5  &    76.5  & \b{97.6} &    77.4  &    72.2  & \b{98.7} &    88.0  &    31.1  &    70.4  &    ~0.2  &    40.7  &    11.8  &    74.3  \\
\ru Corvi et al.  &    79.4  &    73.7  &    50.0  &    97.1  &    63.4  &    65.0  &    80.7  &    91.9  & \g{100.} & \b{100.} & \b{99.9} &    85.7  & \b{100.} & \b{100.} &    69.4  &    60.8  & \b{100.} & \b{98.0} &    84.2  \\
\ru LGrad         & \g{100.} &    91.2  &    83.8  &    81.8  &    82.2  &    80.6  &    76.9  &    66.1  &    81.1  &    61.5  &    68.8  &    74.1  &    56.2  &    57.2  &    58.6  &    37.9  &    56.3  &    40.6  &    69.7  \\
\ru Ojha et al.   & \g{100.} &    93.9  &    92.3  &    98.2  &    96.0  &    58.4  & \b{86.7} &    80.8  &    85.7  &    89.5  &    92.9  &    80.6  &    77.8  &    85.1  & \b{95.2} &    36.4  &    66.2  & \b{97.5} &    84.1  \\
\ru DIRE-1        &    50.6  &    56.9  &    47.8  & \b{99.9} &    74.1  &    44.3  & \g{75.7} &    71.4  &    68.7  &    39.4  & \b{98.9} & \b{99.1} & \b{99.6} &    47.1  &    44.7  &    47.6  &    51.0  &    57.4  &    65.2  \\
\ru DIRE-2        &    54.2  &    52.5  &    43.0  & \b{99.6} &    76.0  &    41.0  &    70.1  &    70.1  &    69.3  &    46.9  &    97.0  &    98.2  &    98.3  &    42.8  &    41.0  &    49.6  &    47.8  &    43.0  &    63.3  \\
\ru NPR           & \g{100.} &    85.6  &    77.0  &    96.4  &    88.7  &    91.1  & \b{86.3} &    79.3  &    90.2  &    64.5  &    91.6  &    80.1  &    78.4  &    76.7  &    39.5  &    48.7  &    77.0  &    32.1  &    76.8  \\ \midrule
\ru Ours 1k       & \b{98.9} &    90.5  &    85.5  & \b{100.} &    81.3  &    89.1  &    81.1  & \b{99.9} & \g{94.1} &    87.6  &    96.5  & \b{98.5} &    94.1  &    87.8  &    89.0  &    70.0  &    73.0  &    74.4  &    88.4  \\
\ru Ours 1k+      &    91.4  &    80.9  &    84.0  & \b{99.8} &    74.7  &    84.3  &    75.2  & \b{99.6} & \g{81.6} &    89.8  &    98.0  & \b{99.1} &    92.5  &    88.9  &    83.6  & \b{93.6} &    78.7  &    85.1  &    87.8  \\
\ru Ours 10k      & \b{99.8} &    91.8  &    86.8  & \b{100.} &    83.6  &    89.0  &    81.4  & \b{99.9} & \g{94.2} &    90.7  &    97.0  & \b{98.7} &    95.0  &    87.4  &    89.2  &    77.6  &    75.3  &    80.1  & \b{89.8} \\
\ru Ours 10k+     &    93.4  &    87.1  &    87.6  & \b{99.9} &    78.5  &    89.2  &    79.9  & \b{99.7} & \g{84.7} &    91.3  &    97.9  & \b{99.4} &    94.0  &    90.1  &    86.3  & \b{92.9} &    81.7  &    87.2  & \b{90.0} \\
 \bottomrule
\end{tabular}
}
\vspace{-1.5mm}
\caption{{\bf Comparison with SoTA methods in terms of AUC.}
For our approach we show four variants:
1k and 10k indicate the number of real and fake training images, + indicates augmentation (compression/resizing).
Results on the dataset used for training are in light gray, while bold underlines the best performance for each dataset with a margin of 1\%.
The last column shows the average over all datasets.}
\label{tab:RES_AUC}
}
\end{table*}

\begin{table*}[t!]
{\small
\centering
\resizebox{1.0\linewidth}{!}{
\begin{tabular}{l |C{7mm}C{7mm}C{7mm}C{7mm}C{7mm}|C{7mm}C{7mm}C{7mm}C{7mm}C{7mm}C{7mm}C{7mm}C{7mm}C{7mm}|C{7mm}C{7mm}C{7mm}C{7mm}|C{7mm}} \toprule
                  &\multicolumn{5}{c|}{GAN family}            & \multicolumn{9}{c|}{Diffusion family}
                  &\multicolumn{4}{c|}{Commercial tools} & AVG \\ \cmidrule(lr){2-6} \cmidrule(lr){7-15} \cmidrule(lr){16-19}
                  &      Pro &    Style &   Style  &   Style  &    Giga  &   Score  &\twr{ADM} &\twr{GLIDE}&  Latent  &  Stable  & DeepFl.  &\twr{Ediff-I}& \twr{DiT}&\twr{SDXL}& DALL· & DALL·  & \twr{Midj.} & Adobe   &    \\[-0mm]
     Method       &      GAN &     GAN2 &    GAN3  &    GANT  &     GAN  &     SDE  &          &          &   Diff.  &   Diff.  &      IF  &          &          &          &    E2    &     E3   &          &   Firef.       &   \\ \midrule
\ru Wang et al.   & \g{100.} &    86.6  &    88.4  &    61.7  &    59.2  &    68.0  &    65.0  &    60.6  &    67.1  &    55.2  &    50.3  &    48.3  &    55.1  &    64.5  &    46.2  &    27.7  &    46.7  &    55.9  &    61.5  \\
\ru PatchFor.     & \g{57.9} &    51.8  &    57.0  &    50.6  &    53.8  &    69.0  &    66.2  &    83.3  &    58.8  &    48.3  &    61.3  &    65.0  &    68.1  &    63.3  &    64.3  &    63.3  &    59.0  &    65.1  &    61.4  \\
\ru Grag. et al.  & \g{100.} & \b{95.4} & \b{90.9} &    94.4  &    64.4  &    77.1  &    77.1  &    79.8  & \b{84.0} &    53.5  &    50.6  &    55.6  &    66.7  &    66.6  &    55.2  &    25.1  &    48.5  &    60.2  &    69.2  \\
\ru Mand. et al.  &    81.1  & \g{79.3} &    87.2  &    49.1  &    49.3  & \g{64.0} &    54.8  &    42.6  &    52.9  &    39.4  &    55.7  &    54.5  &    49.8  &    42.2  &    47.9  &    42.3  &    35.2  &    53.4  &    54.5  \\
\ru Liu et al.    & \g{64.3} &    55.1  &    50.1  &    57.3  &    45.4  &    62.6  &    51.1  &    58.6  &    50.7  &    58.6  &    50.9  &    64.2  &    53.9  &    56.0  &    44.4  &    61.8  &    52.6  &    53.1  &    55.0  \\
\ru Corvi et al.  &    77.5  &    74.7  &    69.4  &    82.1  &    66.6  &    70.4  &    79.0  &    93.5  & \g{99.3} &    69.9  &    60.7  &    72.1  &    89.2  &    61.8  &    65.9  &    32.4  &    51.9  &    58.1  &    70.8  \\
\ru LGrad         & \g{56.3} &    58.3  &    49.8  &    52.3  &    43.5  &    45.9  &    49.2  &    42.3  &    50.4  &    54.8  &    40.7  &    46.4  &    49.4  &    53.2  &    41.8  &    53.5  &    50.4  &    51.8  &    49.4  \\
\ru Ojha et al.   & \g{99.8} &    75.5  &    75.4  &    91.1  & \b{88.5} &    79.3  & \b{83.7} &    83.3  &    81.8  &    75.0  &    59.9  &    68.7  &    70.1  &    61.8  &    63.2  &    41.7  &    40.6  &    52.9  &    71.8  \\
\ru DIRE-1        &    48.4  &    42.5  &    39.1  &    53.5  &    54.3  &    44.1  & \g{48.0} &    44.7  &    46.1  &    47.0  &    66.2  &    62.8  &    53.2  &    47.1  &    44.6  &    47.6  &    51.0  &    57.4  &    49.9  \\
\ru DIRE-2        &    49.3  &    41.6  &    38.6  &    53.8  &    55.0  &    44.3  &    45.1  &    40.2  &    45.9  &    56.4  &    70.7  &    72.2  &    53.0  &    42.8  &    40.9  &    49.7  &    47.8  &    43.0  &    49.5  \\
\ru NPR           & \g{54.5} &    48.5  &    41.9  &    54.0  &    44.8  &    44.7  &    46.9  &    47.2  &    47.7  &    55.4  &    49.6  &    54.6  &    50.9  &    52.8  &    50.0  &    67.5  &    50.8  &    55.5  &    51.0  \\ \midrule
\ru Ours 1k       & \b{85.0} &    64.0  &    66.6  &    90.2  &    75.2  &    74.7  &    78.1  &    97.2  & \g{77.1} &    77.6  &    80.1  &    86.6  &    77.5  &    76.5  &    77.9  &    77.4  &    63.1  &    70.5  &    77.5  \\
\ru Ours 1k+      &    78.7  &    62.5  &    68.4  & \b{97.5} &    67.9  &    84.0  &    74.3  & \b{99.6} & \g{78.2} &    83.7  &    94.5  &    97.1  &    88.9  & \b{89.6} &    81.2  & \b{90.9} &    77.6  &    83.7  &    83.2  \\
\ru Ours 10k      & \b{85.7} &    65.5  &    68.1  &    90.5  &    74.7  &    75.8  &    78.4  &    97.7  & \g{77.8} &    78.1  &    81.2  &    87.1  &    77.2  &    76.4  &    78.2  &    76.4  &    65.0  &    72.2  &    78.1  \\
\ru Ours 10k+     &    82.8  &    67.4  &    70.7  & \b{98.4} &    71.9  & \b{85.4} &    77.3  & \b{99.7} & \g{80.2} & \b{85.8} & \b{95.9} & \b{98.2} & \b{91.1} & \b{89.9} & \b{83.8} & \b{90.1} & \b{79.4} & \b{85.5} & \b{85.2} \\
 \bottomrule
\end{tabular}
}
\vspace{-1.5mm}
\caption{{\bf Comparison with SoTA methods (AUC) on images post-processed by random cropping, resizing and compression.}
}
\label{tab:RES_AUC_PP}
}
\end{table*}

\section{Comparison with the state-of-the-art}
\label{sec:results}

In this Section we perform an extensive comparison with SoTA methods in different scenarios.
We consider four versions of our approach, with 1,000+1,000 (1k) or 10,000+10,000 (10k) real+fake images in the reference set, and with (+) or without compressed/resized images for augmentation.
To ensure a fair comparison, we only include SoTA methods with code and/or pre-trained models publicly available online.
They are listed in Tab.~\ref{tab:sota_methods} and described in the supplementary material.

\vspace{2mm} \noindent
{\bf Generalization analysis.}
In Tab.~\ref{tab:RES_AUC} we show the results in terms of AUC on 18 generative models.
For each method, the results obtained on the same dataset used for training are in light gray, since the ID scenario is of little interest for our analysis.
The items in bold, instead, highlight the best OOD performance for each dataset, considering a margin of 1\%.
We can observe that methods trained on one GAN dataset generally perform well on other datasets of the same family but fare much worse on DM datasets.
Then, the roles change for methods trained on DM datasets.
This is unsurprising, as generators of the same family share architectural details that leave similar traces on the generated images.
No SoTA method performs uniformly well on all datasets.
In contrast, the proposed lightweight CLIP-based detector consistently delivers strong performance.
The version with 10k+10k reference images, without augmentation, outperforms the best competitor by +6.8\% in terms of average AUC.
For GAN-based generators, the proposed CLIP-based detectors keep providing the best performance, generally much better than SoTA methods, except for the Liu method which works almost at the same level.
However, the performance of this method, as well as many other methods, decreases catastrophically when considering synthetic images from commercial tools.
This is the most realistic and interesting scenario, with images of unknown origin and no prior information on the possible generation process.
In this situation, most methods provide unreliable decisions.
Corvi's method works surprisingly well on some datasets, perhaps generated by diffusion models similar to Latent.
However, only the proposed CLIP-based approach provides good stable performance across all cases.
On commercial tools, the versions with resized and recompressed images in the reference set prove especially strong.

\vspace{2mm} \noindent
{\bf Robustness to perturbations.}
The observed trend with respect to unknown models is further accentuated when the images undergo post-processing, as shown in Tab.~\ref{tab:RES_AUC_PP}.
These impairments attenuate forensic traces, to the point that most SoTA methods become essentially useless, performing no better than random chance, particularly on unknown commercial models.
CLIP-based detectors, instead, keep providing a good performance.

\section{Beyond low-level forensic traces}
\label{sec:features}

As previously mentioned, conventional detectors often experience significant performance degradation when images are resized or compressed, as the subtle forensic traces they depend on are significantly reduced.
In contrast, the proposed CLIP-based detector keeps working well under these conditions, suggesting that it relies on higher-level semantic features.
To further investigate this hypothesis, we now conduct some targeted experiments.

\begin{figure}[t!]
	\centering
	\setlength{\tabcolsep}{0.1em}
	\begin{tabular}{cccc}
		\includegraphics[width=2.05cm]{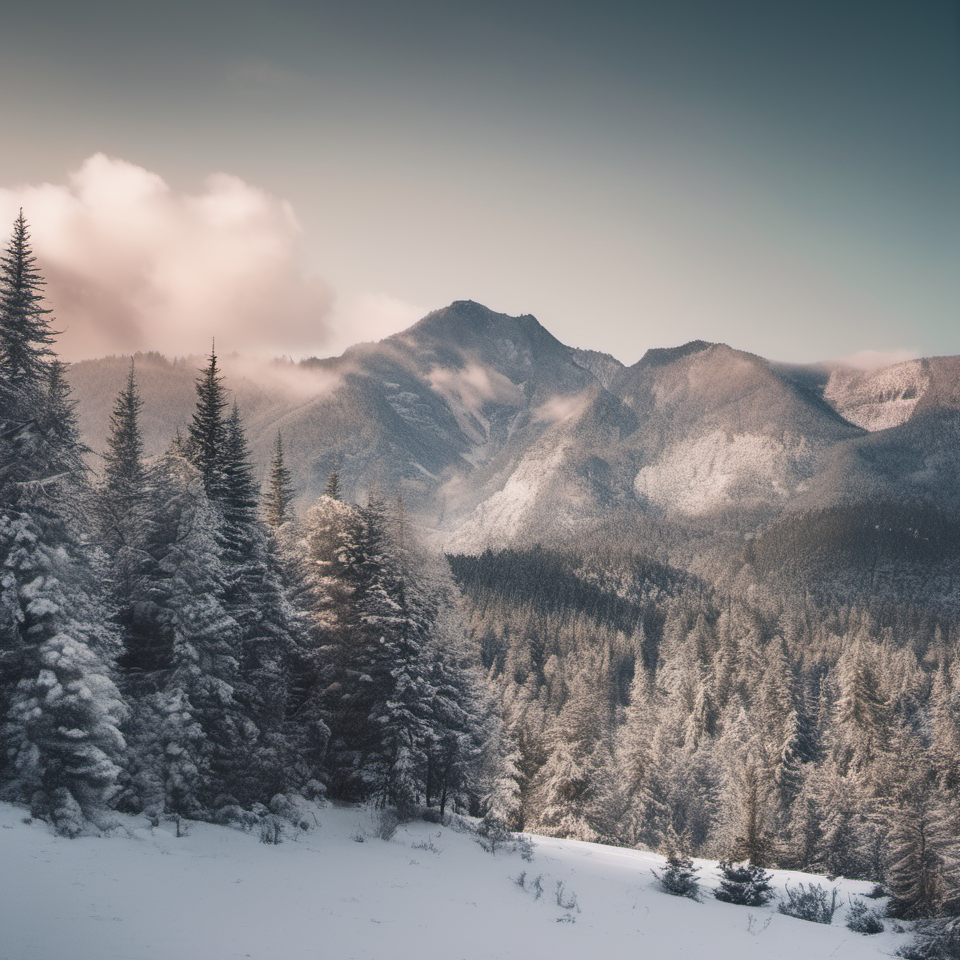} &
		\includegraphics[width=2.05cm]{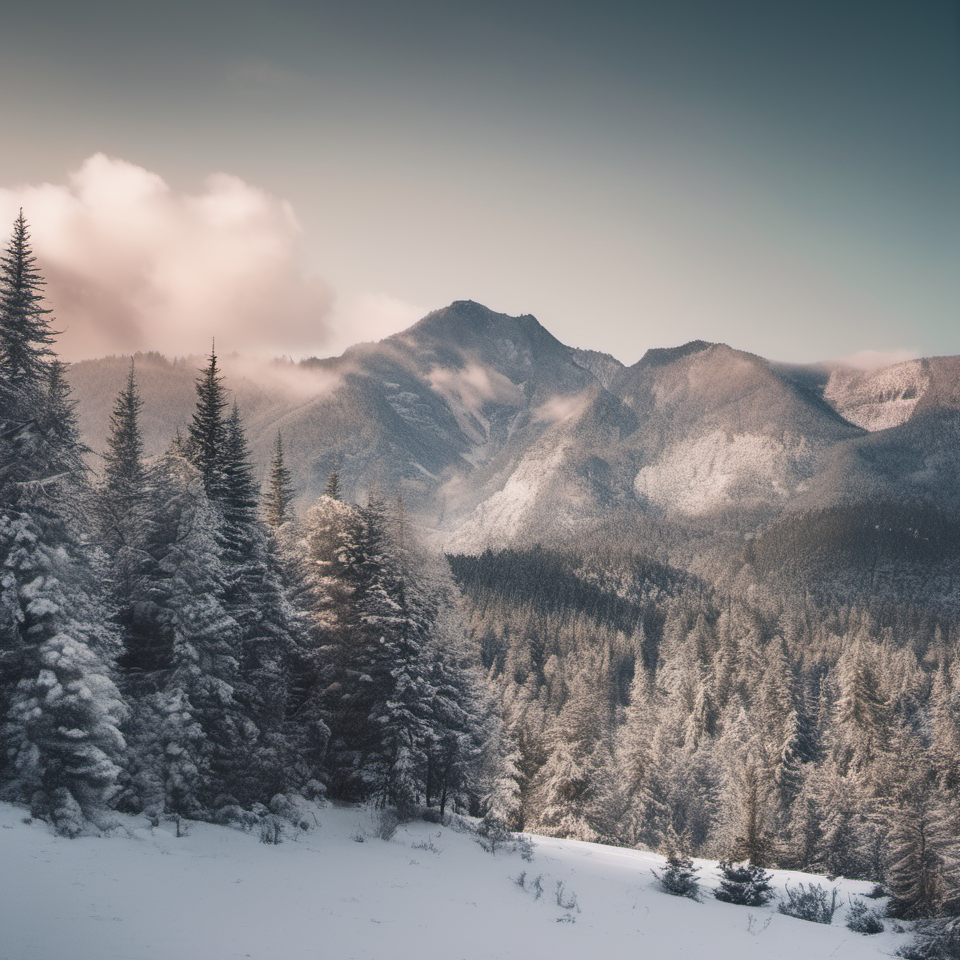} &
		\includegraphics[width=2.05cm]{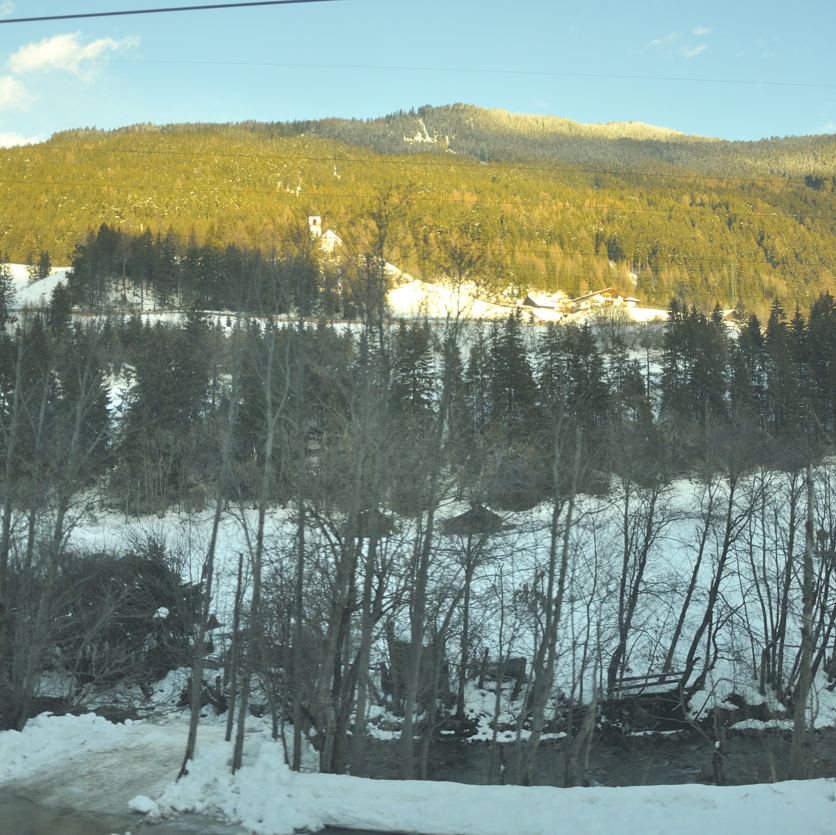} &
		\includegraphics[width=2.05cm]{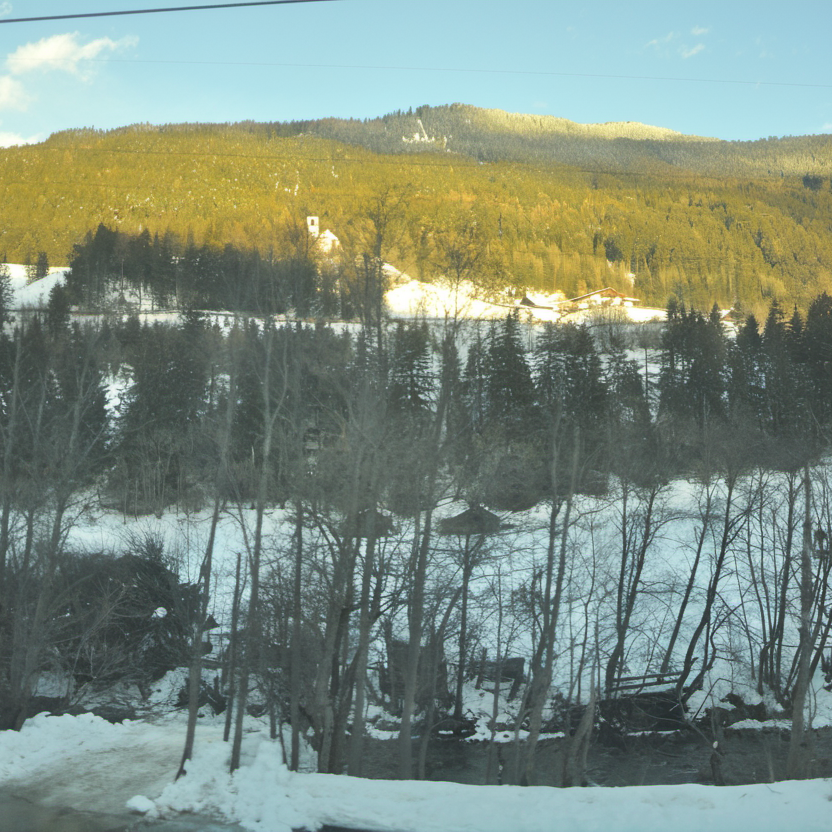}
            \vspace{0.15cm} \\
		\includegraphics[width=2.05cm]{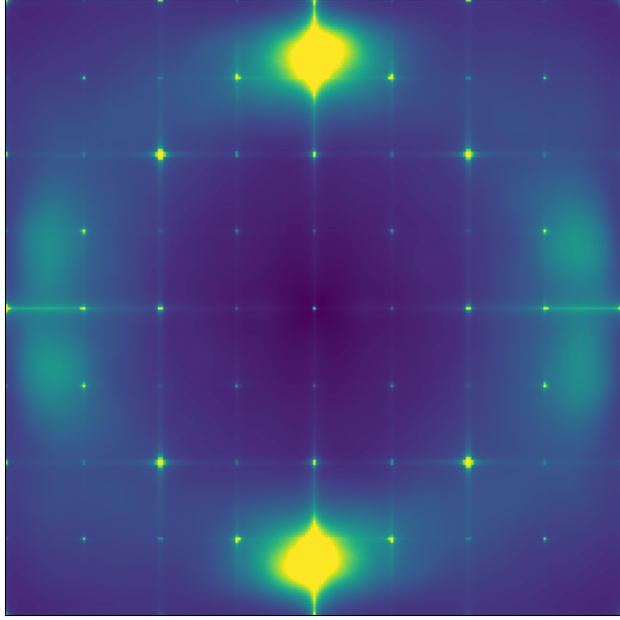} &
		\includegraphics[width=2.05cm]{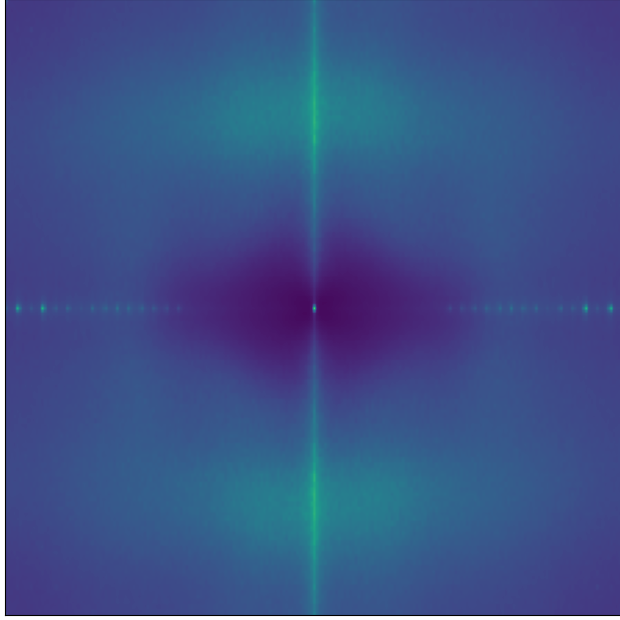} &
		\includegraphics[width=2.05cm]{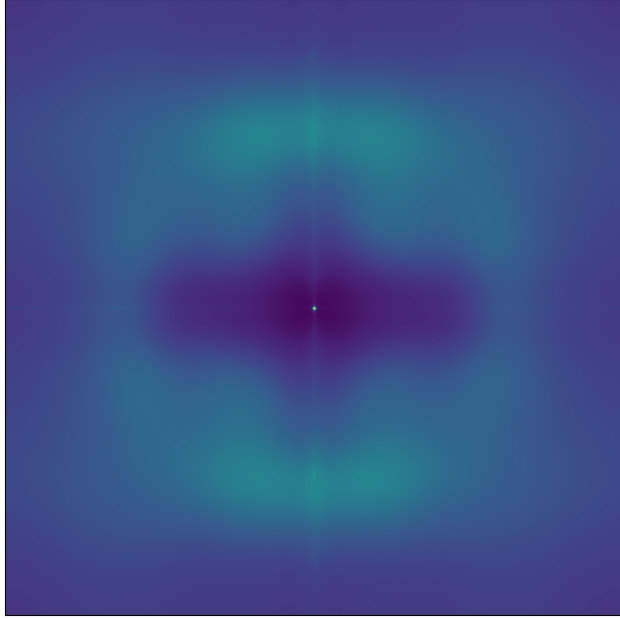} &
		\includegraphics[width=2.05cm]{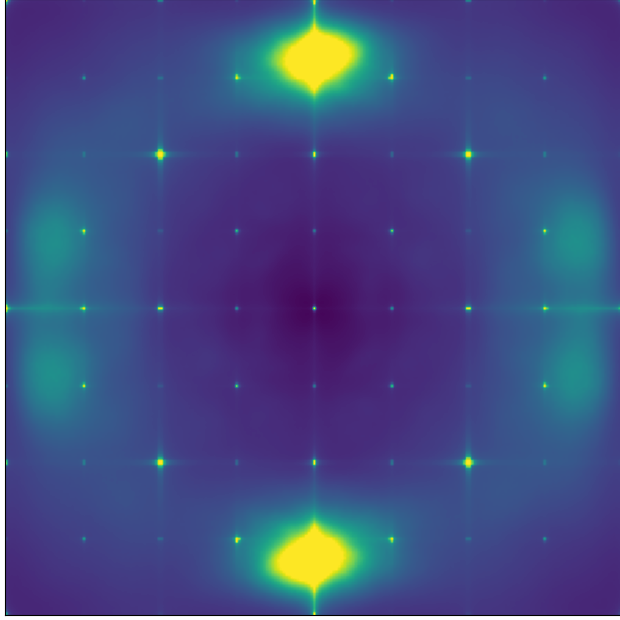}
            \vspace{0.15cm} \\
	\end{tabular}
\caption{Top (from left to right): a synthetic image generated by Stable Diffusion XL \cite{podell2023sdxl} and its  4$\times$ decimated version; a real image and the corresponding image processed by the autoencoder of Stable Diffusion XL. Bottom: Fourier spectra of the noise residuals for images shown on the top. A suitable decimation removes Fourier peaks in synthetic images, while passing a real image through an autoencoder creates new peaks.}
\label{tig:beyond}
\end{figure}

\vspace{2mm} \noindent
{\bf Removing/inserting low-level traces in images.}
We take synthetic images generated by Stable Diffusion 1.4, Stable Diffusion 2.0 \cite{stablediffusion2}, Stable Diffusion XL \cite{podell2023sdxl}) and downsample them by 1/4.
Fig.~\ref{tig:beyond} shows, on the left, the high- and low-resolution images (top) along with their spectra (bottom) for Stable XL.
After reducing the resolution, the peaks in the Fourier domain completely disappeared, because they were caused by the aliasing effect of the oversampling filters, which can be removed by appropriate decimation.
In a complementary experiment, we take real images from the RAISE dataset and pass them through the autoencoders used by Stable Diffusion XL.
This process does not change the visual appearance of the images (Fig.~\ref{tig:beyond} top-right) but inserts low-level forensic traces in their spectra (bottom) that resemble those of the generated images.

In Table \ref{tab:autoencoder} we quantify the effects of such attacks on 1,000 real and 1,000 synthetic images, reporting the AUC for our method and the best competitor before the attack (first group of columns) after removing traces from synthetic images (second group) and after adding them to real images (third group).
Before the attack, all detectors work quite well, with the method proposed by Corvi et al., based on low-level features, reaching perfect discrimination.
After the attacks, however, the performance of Corvi et al. degrades dramatically, while the proposed method suffers only a very limited loss.
In summary, the CLIP-based detector appears to withstand various types of image degradations easily, both innocent and malicious.
The latter case is particularly relevant: attacks can be carried out with the very aim of making everything appear as synthetic, misleading current forensic detectors based on low-level traces, and undermining public trust in forensic analyses.

\begin{figure}
    \centering
    \includegraphics[height=0.49\linewidth,page=1,clip,trim=0 0 60 0]{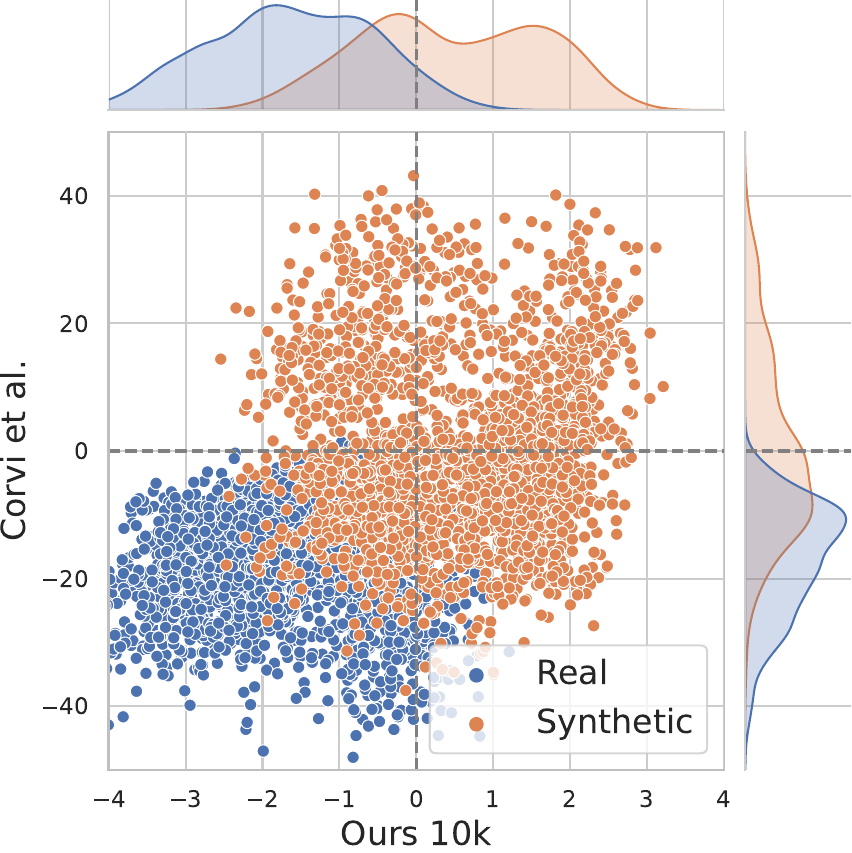}
    \includegraphics[height=0.49\linewidth,page=2,clip,trim=15 0 0 0]{figures/scatter_feature.pdf}
    \caption{{\bf Scatter plots} of scores provided by Corvi et al. ($y$-axis) and proposed method without and with augmentation ($x$-axis) on 2400+2400 images sampled from the 18 datasets of Tab.\ref{tab:datasets}.}
\label{fig:scatter}
\end{figure}

\vspace{2mm} \noindent
{\bf Fusion.}
Based on the above results, the proposed CLIP-based detector appears not to rely on the same low-level traces exploited by most of the current detectors.
This is further supported by the scatter plot of Fig.\ref{fig:scatter} where each point gives the scores of the CLIP-based and Corvi detectors for a given image.
The two sets of scores are almost orthogonal, as if they depended on uncorrelated features, a property that paves the way for appropriate fusion strategies.
We implemented a simple decision rule where the image is declared real only if both detectors agree on this choice.
Tab.\ref{tab:fusione} reports AUC and Accuracy results for Corvi, the proposed method (10k and 10k+ versions), and their fusion. Results are given as averages over the three families of generators: GAN, Diffusion and Commercial Tools.
The fusion ensures a further boost in performance over the proposed CLIP-based detector, both in terms of AUC (+3.6\%) and Accuracy (+7.4\%). 
A smaller improvement is observed when images are resized/compressed, arguably because low-level traces are more compromised in this case.

\section{Discussion}
\label{sec:conclusions}

We have proposed a simple method based on CLIP features to distinguish real images from synthetic images. Through extensive experimental analysis, we found that:
\begin{itemize}
\item   CLIP features support a much higher generalization ability than what had been discovered so far. 
        By leveraging just a few examples, not even belonging to the generator under test, a simple CLIP-based detector achieves top performance on a large variety of generators and in the most challenging conditions.
\item   Maximizing the diversity of reference CLIP features has a positive impact on performance, even when a relatively large number of examples is considered.
\item   It is known that CLIP's descriptive power comes from its huge pre-training set. Further increasing this set keeps boosting performance, up to $10\%$.
\item   Experimental evidence suggests that CLIP features, even when adapted to forensic applications, are largely independent of low-level forensic traces. 
        This provides some immunity to malicious attacks targeting low-level artifacts and paves the way for further performance improvements through suitable fusion with traditional detectors.
\end{itemize}
The present study provides a number of interesting insights but leaves much room for future work.
We believe that new and better forensic methods can be proposed based on CLIP features.
One major area of study is the development of few-shot methods that adapt the detector to the situation of interest on the fly.
Future work should also consider interpretability, starting with understanding which forensic features the detector exploits to make its decisions.

\begin{table}[t!]
{\small
\centering
\resizebox{1.\linewidth}{!}{
\setlength{\tabcolsep}{0.5pt}
\begin{tabular}{l| C{11mm}C{11mm}C{11mm} | C{11mm}C{11mm}C{11mm} | C{11mm}C{11mm}C{11mm}} \toprule
 \ru  &    \multicolumn{3}{c|}{}  & \multicolumn{3}{c|}{downsample by 1/4} & \multicolumn{3}{c}{add low-level traces}              \\ \cmidrule(lr){2-4} \cmidrule(lr){5-7} \cmidrule(lr){8-10}
 \ru \twr{Method}          &  St.Diff.  & St.Diff.   &    \twr{SDXL}   &   St.Diff. &   St.Diff.  &    \twr{SDXL} &   St.Diff. &   St.Diff.  &    \twr{SDXL}      \\[-0mm]
 \ru         &  1.4 &  2.0  &     &   1.4 &  2.0  &     &   1.4 &  2.0  &       \\
 \midrule
\ru Corvi et al.      &  100. &  100. &  100. &  41.6 &  46.2 &  45.6 &  79.3 &  80.4 &  67.6 \\
\ru Ours 1k           &  88.8 &  87.0 &  87.8 &  85.8 &  82.2 &  79.3 &  89.3 &  82.5 &  80.8 \\
\ru Ours 1k+          &  90.2 &  89.4 &  88.9 &  89.8 &  91.1 &  90.0 &  90.8 &  86.1 &  83.2 \\
\ru Ours 10k          &  93.8 &  90.6 &  87.4 &  86.8 &  84.4 &  80.5 &  93.7 &  86.2 &  80.6 \\
\ru Ours 10k+         &  94.0 &  90.4 &  90.1 &  95.9 &  92.6 &  91.6 &  94.1 &  86.7 &  84.0 \\
\bottomrule
\end{tabular}
}
\vspace{-1.5mm}
\caption{Results in terms of AUC without attacks (first group of columns)
and after attacks on synthetic images (second group) and real images (third group).
Corvi et al., based on low-level forensic traces, is severely affected by the attacks
while the proposed method keeps working well in all conditions.}
\label{tab:autoencoder}
}
\end{table}

\begin{table}[t!]
{\small
\centering
\resizebox{1.\linewidth}{!}{
\begin{tabular}{l| C{18mm}C{18mm}C{18mm} |C{20mm}} \toprule
 \ru              &    \multicolumn{3}{c|}{Families of Generators}            &                   \\ \cmidrule(lr){2-4}
 \ru              &    GAN            &    Diffusion      &    Comm. Tools    &    Average        \\
 \ru Method       &    AUC/Acc        &    AUC/Acc        &    AUC/Acc        &    AUC/Acc        \\ \midrule
\ru Corvi et al.  &    72.7  /  52.1  &    91.5  /  75.1  &    82.1  /  62.8  &    82.1  /  63.3  \\
\ru Ours 10k      &    92.4  /  79.1  &    92.6  /  73.3  &    80.5  /  52.6  &    88.5  /  68.3  \\
\ru Ours 10k+     &    89.3  /  74.9  &    91.8  /  77.2  &    87.0  /  67.3  &    89.4  /  73.1  \\
\ru Ours fusion   &    92.9  /  80.1  &    96.9  /  87.8  &    88.2  /  65.3  &    92.7  /  77.8  \\
\ru Ours fusion+  &    89.9  /  75.5  &    96.8  /  88.8  &    92.3  /  77.2  &    93.0  /  80.5  \\ \midrule
\ru Corvi et al.  &    74.0  /  55.1  &    77.3  /  62.1  &    52.1  /  50.1  &    67.8  /  55.8  \\
\ru Ours 10k      &    76.9  /  63.6  &    81.1  /  63.7  &    73.0  /  52.4  &    77.0  /  59.9  \\
\ru Ours 10k+     &    78.2  /  69.0  &    89.3  /  78.7  &    84.7  /  66.4  &    84.1  /  71.4  \\
\ru Ours fusion   &    78.5  /  66.8  &    85.1  /  71.4  &    72.7  /  52.6  &    78.7  /  63.6  \\
\ru Ours fusion+  &    79.6  /  70.8  &    91.7  /  80.7  &    84.5  /  66.6  &    85.3  /  72.7  \\
 \bottomrule
\end{tabular}
}
\vspace{-1.5mm}
\caption{{\bf AUC/Accuracy}
results for Corvi et al., proposed method (10k and 10k+ versions), and their fusion
over the three families of generators: GAN, Diffusion, Commercial Tools.
Top: original images; bottom: compressed/resized images.}
\label{tab:fusione}
}
\end{table}

\paragraph{Acknowledgment.}
{
We gratefully acknowledge the support of this research by a TUM-IAS Hans Fischer Senior Fellowship, a TUM-IAS Rudolf M\"o{\ss}bauer Fellowship and a Google Gift.
This material is also based on research sponsored by the Defense Advanced Research Projects Agency (DARPA) and the Air Force Research Laboratory (AFRL) under agreement number FA8750-20-2-1004.
In addition, this work has received funding by the European Union under the Horizon Europe vera.ai project, Grant Agreement number 101070093.
Finally, we want to thank Koki Nagano, Tero Karras, Yogesh Balaji and Ming-Yu Liu for sharing data for StyleGAN-T and eDiff-I experiments.
}

\begin{appendix}
\section*{Supplemental Material}
In this appendix,
we provide a brief description of the methods used for comparison (Sec. \ref{sota_methods}),
report additional ablation studies (Sec. \ref{ablation}), 
additional results (Sec. \ref{accresults}), 
and robustness analysis (Sec. \ref{robustness}).
We also show some experiments carried out on social networks in a few-shot scenario (Sec. \ref{few-shot}).
Finally, we enlarge our initial dataset with additional synthetic generators and carry out further experiments on generalization (Sec. \ref{generalization}). 

\renewcommand{\b}[1]{{\bf #1}}

\section{Reference methods}
\label{sota_methods}

In our experimental analysis we included the following methods:

\begin{enumerate}[label=(\alph*)]
    \item {\bf Wang~et~al.}~\cite{wang2020cnn} is a CNN detector based on ResNet50 and represents a reference in the research community. This work also introduced the large dataset (LSUN/ProGAN) extensively adopted for model training in subsequent works.
    \item {\bf Gragnaniello~et~al.}~\cite{gragnaniello2021GAN} proposes a simple modification to the ResNet50 architecture which allows to better preserve low-level forensic traces and is trained on the same dataset introduced in \cite{wang2020cnn}.
    \item {\bf Mandelli et al.}~\cite{mandelli2022detecting} relies on the ensemble of five EfficientNet-B4 networks trained on different datasets. At test-time the scores of randomly selected patches are aggregated: if at least one patch is detected as synthetic, then the entire image is classified as synthetic.
    \item {\bf PatchForensics}~\cite{chai2020what} develops a fully-convolutional classifier based on local patches with limited receptive fields over an XceptionNet backbone.
    \item {\bf Liu et al.}~\cite{liu2022detecting} is a detector that exploits the inconsistency between real and fake images in the representations of the learned noise patterns. This spatial information is combined with frequency information to improve the classification.
    \item {\bf LGrad}~\cite{tan2023learning} works on the gradients extracted through a pre-trained CNN model in order to filter out the content of the image and transform a data-dependent problem into a transformation-model dependent problem.
    \item {\bf Corvi et al.}~\cite{corvi2022detection} performs strong augmentation to gain robustness and increase generalization and is trained on a large dataset of latent diffusion models.
    \item {\bf Ojha et al.}~\cite{ojha2023towards} uses a large dataset of real and synthetic images to train a simple classifier working on pre-trained CLIP features.
    \item {\bf DIRE}~\cite{wang2023dire} is based on the idea that synthetic images can be reconstructed better than real images by a pre-trained model. To this end a ResNet-50 is trained in two different ways, using ADM images (DIRE 1) and StyleGAN images (DIRE 2), respectively.
    \item {\bf NPR}~\cite{tan2023rethinking} works on residual images computed as the difference between the original image and its interpolated version. The classifier is a ResNet-50  trained on only 4 classes of the ProGAN dataset.
\end{enumerate}

\begin{figure}[t!]
	\centering
   \includegraphics[page=1, width=0.45\linewidth, trim=0 0 0 0,clip]{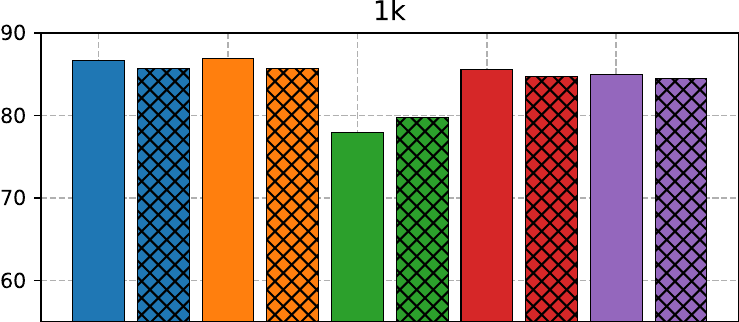} 
   \includegraphics[page=2, width=0.45\linewidth, trim=0 0 0 0,clip]{figures/classAUC.pdf} \\[1mm] 
   \includegraphics[page=3, width=0.45\linewidth, trim=0 0 0 0,clip]{figures/classAUC.pdf} 
   \includegraphics[page=4, width=0.45\linewidth, trim=0 0 0 0,clip]{figures/classAUC.pdf} \\
   \includegraphics[page=5, width=1.00\linewidth, trim=0 5 0 0,clip]{figures/classAUC.pdf} \\
	\caption{Average AUC for our approach 
 considering different classifiers and different features on the original dataset (top) and after random compression and resizing (bottom). Left: 1k images, Right: 10k images.}
	\label{fig:classifier_ex}
\end{figure}

\section{Additional ablation study}
\label{ablation}

In Section 4 of the paper we describe the CLIP-based detector used in all experiments.
In the proposed procedure two key design choices are made:
{\it  i)} we extract features from the next to last layer of the CLIP ViT L/14 architecture; and
{\it ii)} we use a SVM classifier, trained on a limited set of reference features.
Here we test alternative solutions, that is:
\begin{itemize}
\item   extracting features from the last layer of the architecture;
\item   using other classifiers: Logistic Regression (LR), Mahalanobis distance (MAH), Gaussian Naive Bayes classifier (GNB), Soft voting k-Nearest Neighbor (SNN) \cite{mitchell2001soft}.
\end{itemize}
In Figure \ref{fig:classifier_ex}, we show the results in terms of average AUC over the dataset described in Section 3 (main paper)
using both original images and images impaired by common post-processing steps such as recompression and resizing.
For all detectors, we consider the versions 1k and 10k (1000 and 10000 reference images per class, respectively).

In all cases, the SVM classifier seems to ensure the best performance, followed closely by the Logistic Regression, while the other classifiers provide a less consistent performance. 
Using features from the next-to-last layer is almost always preferable to using features from the last layer,
and this happens always with the SVM classifier.
These results motivated our design choices.

\begin{table}[t!]
{\small
\centering
\resizebox{1.\linewidth}{!}{
\begin{tabular}{l| C{18mm}C{18mm}C{18mm} |C{20mm}} \toprule
 \ru              &    \multicolumn{3}{|c|}{Families of Generators}                 &                    \\ \cmidrule(lr){2-4}
 \ru              &    GAN              &    Diffusion        &    Comm. Tools      &    Average         \\
 \ru Method       &    AUC/Acc          &    AUC/Acc          &    AUC/Acc          &    AUC/Acc         \\ \midrule
\ru Wang et al.       &    92.1  /    69.4  &    55.8  /    50.3  &    50.2  /    49.8  &    66.0  /    56.5  \\
\ru PatchFor.         &    84.9  /    53.6  &    76.4  /    50.4  &    50.5  /    50.0  &    70.6  /    51.3  \\
\ru Grag. et al.      &    95.8  / \b{90.1} &    70.8  /    57.0  &    41.8  /    43.6  &    69.5  /    63.6  \\
\ru Mand. et al.      &    88.9  /    81.6  &    55.8  /    53.7  &    22.6  /    32.9  &    55.7  /    56.1  \\
\ru Liu et al.        & \b{99.0} / \b{89.7} &    79.9  /    69.5  &    30.8  /    42.9  &    69.9  /    67.4  \\
\ru Corvi et al.      &    72.7  /    52.1  &    91.5  /    75.1  &    82.1  /    62.8  &    82.1  /    63.3  \\
\ru LGrad             &    87.8  /    76.4  &    69.2  /    60.4  &    48.3  /    52.1  &    68.4  /    63.0  \\
\ru Ojha et al.       &    96.1  /    85.4  &    82.0  /    63.4  &    73.8  /    68.3  &    84.0  /    72.4  \\
\ru DIRE-1            &    65.9  /    64.4  &    71.6  /    72.8  &    50.2  /    49.9  &    62.5  /    62.4  \\
\ru DIRE-2            &    65.1  /    60.1  &    70.4  /    65.8  &    45.3  /    49.9  &    60.3  /    58.6  \\
\ru NPR               &    89.5  /    79.7  &    82.0  /    68.1  &    49.3  /    50.1  &    73.6  /    66.0  \\ \midrule
\ru Ours 1k           &    91.2  /    76.4  & \b{92.1} /    76.2  &    76.6  /    54.5  &    86.6  /    69.0  \\ 
\ru Ours 1k+          &    86.2  /    74.2  &    89.9  / \b{79.9} &    85.3  / \b{72.6} &    87.1  / \b{75.6} \\
\ru Ours 10k          &    92.4  /    79.1  & \b{92.6} /    73.3  &    80.5  /    52.6  & \b{88.5} /    68.3  \\
\ru Ours 10k+         &    89.3  /    74.9  & \b{91.8} /    77.2  & \b{87.0} /    67.3  & \b{89.4} /    73.1  \\
 \bottomrule
\end{tabular}
}
\vspace{-1.5mm}
\caption{{\bf Comparison with SOTA methods in terms of AUC and Accuracy.} 
We report the mean AUC and Accuracy for each family of generators and the global average.}
\label{tab:WithoutPostProcessing}
}
\end{table}

\begin{table}[t!]
{\small
\centering
\resizebox{1.\linewidth}{!}{
\begin{tabular}{l| C{18mm}C{18mm}C{18mm} |C{20mm}} \toprule
 \ru              &    \multicolumn{3}{|c|}{Families of Generators}                 &                    \\ \cmidrule(lr){2-4}
 \ru              &    GAN              &    Diffusion        &    Comm. Tools      &    Average         \\
 \ru Method       &    AUC/Acc          &    AUC/Acc          &    AUC/Acc          &    AUC/Acc         \\ \midrule
\ru Wang et al.       &    79.2  /    62.0  &    59.3  /    50.4  &    44.1  /    49.9  &    60.9  /    54.1  \\
\ru PatchFor.         &    54.2  /    50.0  &    64.8  /    50.3  &    62.9  /    50.1  &    60.7  /    50.1  \\
\ru Grag. et al.      & \b{89.0} /    67.6  &    67.9  /    50.8  &    47.3  /    50.0  &    68.1  /    56.1  \\
\ru Mand. et al.      &    69.2  /    55.1  &    50.6  /    50.8  &    44.7  /    49.6  &    54.9  /    51.9  \\
\ru Liu et al.        &    54.4  /    51.5  &    56.3  /    51.1  &    53.0  /    50.7  &    54.6  /    51.1  \\
\ru Corvi et al.      &    74.0  /    55.1  &    77.3  /    62.1  &    52.1  /    50.1  &    67.8  /    55.8  \\
\ru LGrad             &    52.0  /    50.9  &    48.0  /    50.5  &    49.4  /    50.4  &    49.8  /    50.6  \\
\ru Ojha et al.       &    86.1  / \b{73.7} &    73.7  /    59.8  &    49.6  /    50.6  &    69.8  /    61.4  \\
\ru DIRE-1            &    47.6  /    50.2  &    51.0  /    50.5  &    50.1  /    49.9  &    49.6  /    50.2  \\
\ru DIRE-2            &    47.7  /    49.6  &    52.3  /    52.5  &    45.3  /    49.9  &    48.4  /    50.7  \\
\ru NPR               &    48.7  /    50.2  &    50.0  /    49.8  &    56.0  /    50.1  &    51.6  /    50.0  \\ \midrule
\ru Ours 1k           &    76.2  /    68.9  &    80.6  /    71.2  &    72.2  /    60.7  &    76.3  /    66.9  \\
\ru Ours 1k+          &    75.0  /    67.2  &    87.8  /    77.8  &    83.4  /    65.7  &    82.1  /    70.3  \\
\ru Ours 10k          &    76.9  /    63.6  &    81.1  /    63.7  &    73.0  /    52.4  &    77.0  /    59.9  \\
\ru Ours 10k+         &    78.2  /    69.0  & \b{89.3} / \b{78.7} & \b{84.7} / \b{66.4} & \b{84.1} / \b{71.4} \\
 \bottomrule
\end{tabular}
}
\vspace{-1.5mm}
\caption{{\bf AUC/Accuracy in the presence of post-processing.} 
All images have been randomly cropped, resized and compressed to simulate a realistic scenario on the web.}
\label{tab:PostProcessing}
}
\end{table}

\section{AUC vs Accuracy}
\label{accresults}

In this Section we present additional results on the synthetic generators analyzed in the main paper aggregated by family (GAN, Diffusion, Commercial tools). 
Results are not only in terms of AUC but also of balanced accuracy computed with a fixed threshold of 0.5. Indeed in a practical scenario a threshold must be set to discriminate between real and fake images, which is not a trivial task as can be seen in Table \ref{tab:WithoutPostProcessing}. In fact, in the absence of an adequate calibration procedure, a significant drop between AUC and accuracy is often observed. However, our approach provides good results even in this challenging context, especially on post-processed images (Tab. \ref{tab:PostProcessing}) where most methods are equivalent to flipping coins.
 
\begin{figure}[t!]
    \centering
    \includegraphics[width=0.32\linewidth, page=1, trim=11 0 0 0,clip]{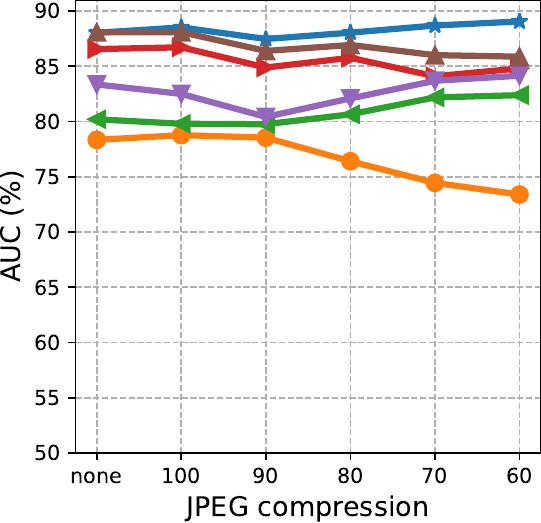}
    \includegraphics[width=0.32\linewidth, page=2, trim=11 0 0 0,clip]{figures/results_robustness.pdf}
    \includegraphics[width=0.32\linewidth, page=3, trim=11 0 0 0,clip]{figures/results_robustness.pdf} 
    \\[1mm]
    \includegraphics[width=1.0\linewidth, page=4]{figures/results_robustness.pdf} 
    \caption{Robustness analysis in terms of AUC carried out on the Stable Diff. 2, SDXL, DALL·E 3, DALL·E 2, Midjourney, and Firefly generation models of the SynthBuster dataset \cite{bammey2023synthbuster}.}
    \label{fig:robustness}
\end{figure}

\section{Additional robustness analysis}
\label{robustness}

In Section 5 of the main paper we provide some results on the robustness of SoTA and proposed methods in the presence of image impairments.
Due to lack of space, however, we show only some very synthetic results, averaged on all kinds of post-processed images.
In Fig. \ref{fig:robustness}, we analyze robustness in more detail as a function of JPEG Quality Factor, ranging from 100 to 60, and resizing scale, going from 125\% to 25\%.
In addition, we consider also another compression method, WebP, which is gaining popularity on social networks but has been rarely considered in experimental analyses.
Images are from the SynthBuster dataset \cite{bammey2023synthbuster}, generated by Stable Diffusion 2, SDXL, DALL·E 2, DALL·E 3, Midjourney, and Firefly.
We show results (AUC) only for the SoTA methods that proved most competitive in terms of robustness, Corvi and Ojha.
For our method we consider again the versions with 1k or 10k real and fake reference images, and the corresponding versions, 1k+ and 10k+, with the same number of images including random  compression and resizing.

Ojha suffers some performance losses, up to 10 points, in the presence of compression, be it JPEG or WebP,
and a much more rapid decline with resizing, to the point of becoming useless in the presence of a 25\% rescaling.
Corvi was trained with strong augmentation and, in fact,
is not affected at all by JPEG compression, while it presents acceptable losses with strong WebP compression (not seen in training), and strong resizing.
The most remarkable outcome of this experiment, however, is the impressive robustness of the CLIP-based detectors.
Both the 1k+ and 10k+ versions, those with augmentation, are basically insensitive to compression, no matter JPEG or WebP, and resizing.
The versions without augmentation suffer some loss of performance but not nearly as dramatic as for the reference methods, and remain effective in all conditions.

\begin{table}[t!]
{\small
\centering
\resizebox{1.\linewidth}{!}{
\begin{tabular}{l| C{14mm}C{14mm}C{14mm} | C{20mm}} \toprule
 \ru AUC/Acc      &    \multicolumn{3}{c|}{Social network: X}                                             &                     \\ \cmidrule(lr){2-4}
 \ru Method       &    DALL·E3          &    Midjourney       &    FireFly          &    Average          \\
 \midrule
\ru Wang et al.   &    22.5  /    50.0  &    36.7  /    50.0  &    65.3  /    50.0  &    41.5  /    50.0  \\
\ru PatchFor.     &    54.5  /    50.0  &    59.9  /    50.0  &    46.5  /    50.0  &    53.6  /    50.0  \\
\ru Grag. et al.  &    44.7  /    49.6  &    58.4  /    49.7  &    80.5  /    49.9  &    61.2  /    49.7  \\
\ru Mand. et al.  &    42.9  /    49.3  &    53.7  /    50.6  &    52.7  /    50.1  &    49.8  /    50.0  \\
\ru Liu et al.    &    34.5  /    49.7  &    38.7  /    49.7  &    54.9  /    49.9  &    42.7  /    49.8  \\
\ru Corvi et al.  &    89.3  /    86.8  &    98.3  /    90.8  &    79.1  /    52.1  &    88.9  /    76.6  \\
\ru LGrad         &    33.7  /    39.0  &    37.9  /    41.4  &    61.4  /    58.4  &    44.3  /    46.3  \\
\ru Ojha et al.   &    34.4  /    45.6  &    32.2  /    46.2  &    73.0  /    65.7  &    46.5  /    52.5  \\
\ru DIRE-1        &    54.3  /    52.0  &    60.7  /    54.3  &    55.1  /    51.1  &    56.7  /    52.5  \\
\ru DIRE-2        &    60.0  /    55.0  &    58.6  /    53.9  &    47.7  /    46.5  &    55.4  /    51.8  \\
\ru NPR           &    43.8  /    41.7  &    46.2  /    44.9  &    65.4  /    62.0  &    51.8  /    49.5  \\  \midrule
\ru Ours 1k       &    69.4  /    50.9  &    70.2  /    53.3  &    69.3  /    51.4  &    69.6  /    51.9  \\
\ru Ours 1k+      &    82.1  /    73.3  &    82.8  /    74.5  &    83.8  /    74.9  &    82.9  /    74.3  \\
\ru Ours 10k      &    65.8  /    50.0  &    69.3  /    51.7  &    70.7  /    50.9  &    68.6  /    50.9  \\
\ru Ours 10k+     &    78.7  /    63.7  &    82.2  /    69.6  &    82.7  /    69.6  &    81.2  /    67.6  \\  \midrule
\ru Ours fusion   &    90.0  /    85.9  &    97.5  /    93.8  &    77.8  /    57.0  &    88.4  /    78.9  \\
\ru Ours fusion+  &    94.7  /    87.3  &    97.9  /    92.1  &    85.4  /    74.2  &    92.6  /    84.5 \\  \midrule
\ru Ours Few-Shot &    98.5  /    92.5  &    95.8 /     87.4  &    92.5 /     83.5  &    95.6  /    87.8  \\
\bottomrule
\end{tabular}
}
\vspace{-1.5mm}
\caption{Results in terms of AUC and accuracy of proposal and SOTA methods on real and synthetic images download from X. The last row reports a few-shot experiment, where we suppose to know in advance 10+10 real/fake images from a specific generator.}
\label{tab:Twitter}
}
\end{table}

\begin{figure}[t!]
	\centering
	\setlength{\tabcolsep}{0.2em}
	\begin{tabular}{cccc}
  \includegraphics[width=2.0cm]{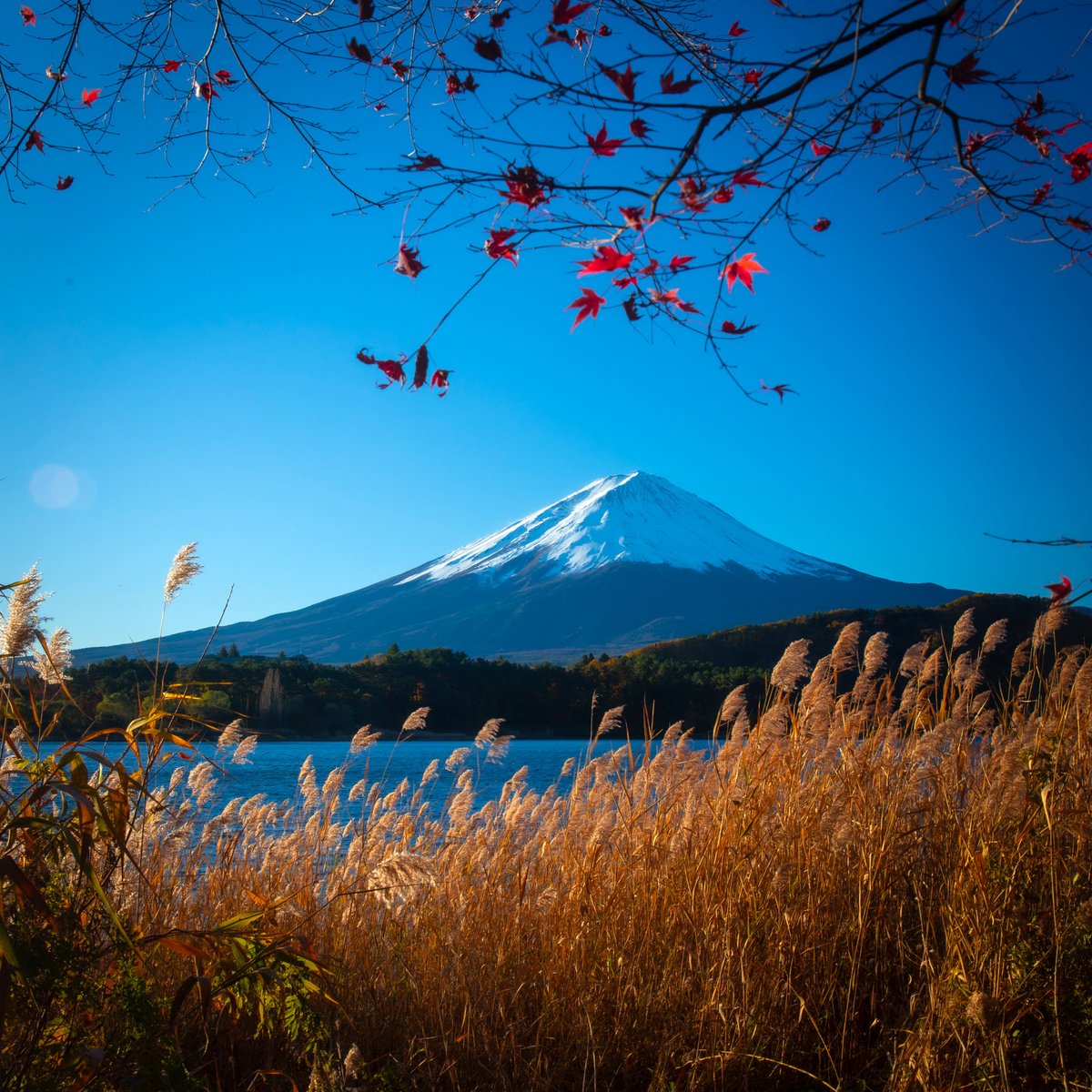} &
  \includegraphics[width=2.0cm]{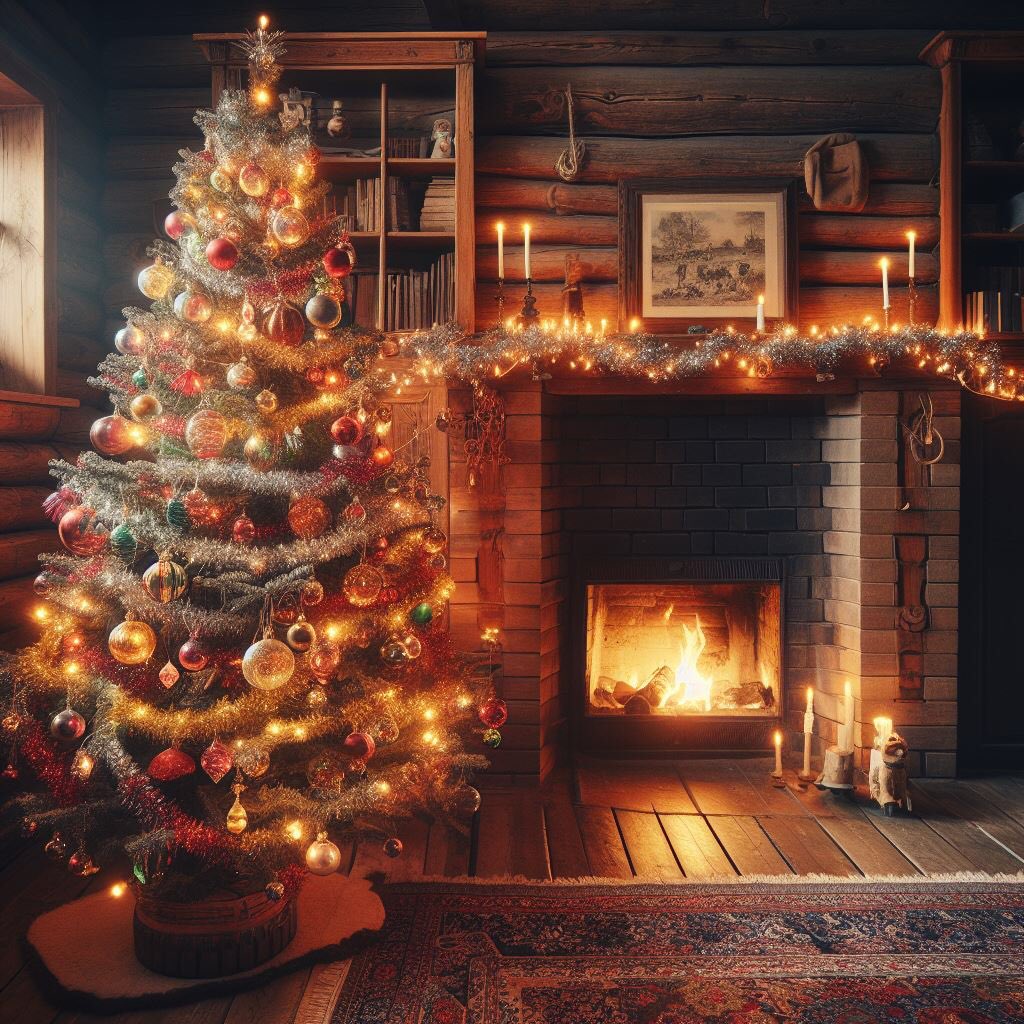} &
  \includegraphics[width=2.0cm]{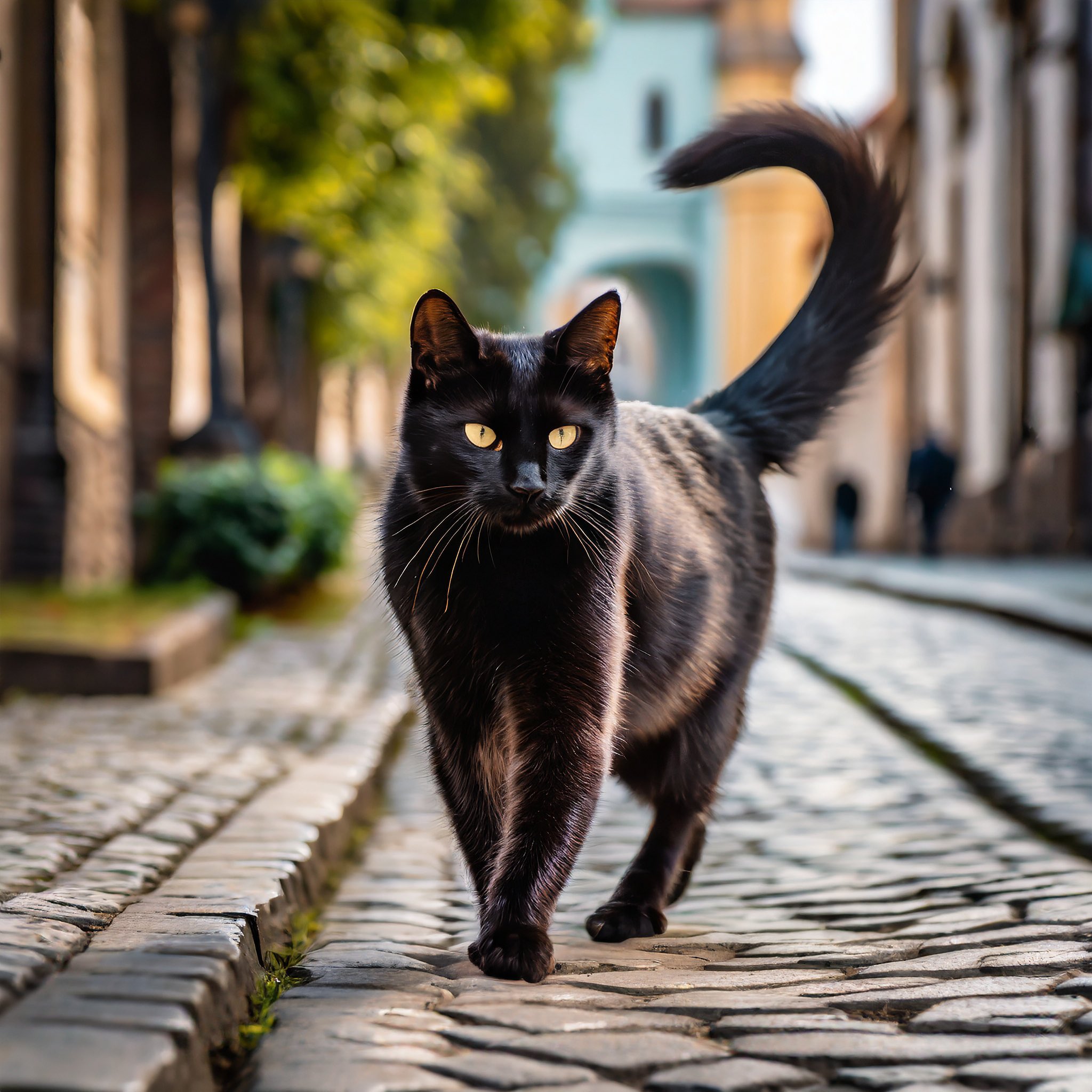}&
  \includegraphics[width=2.0cm]{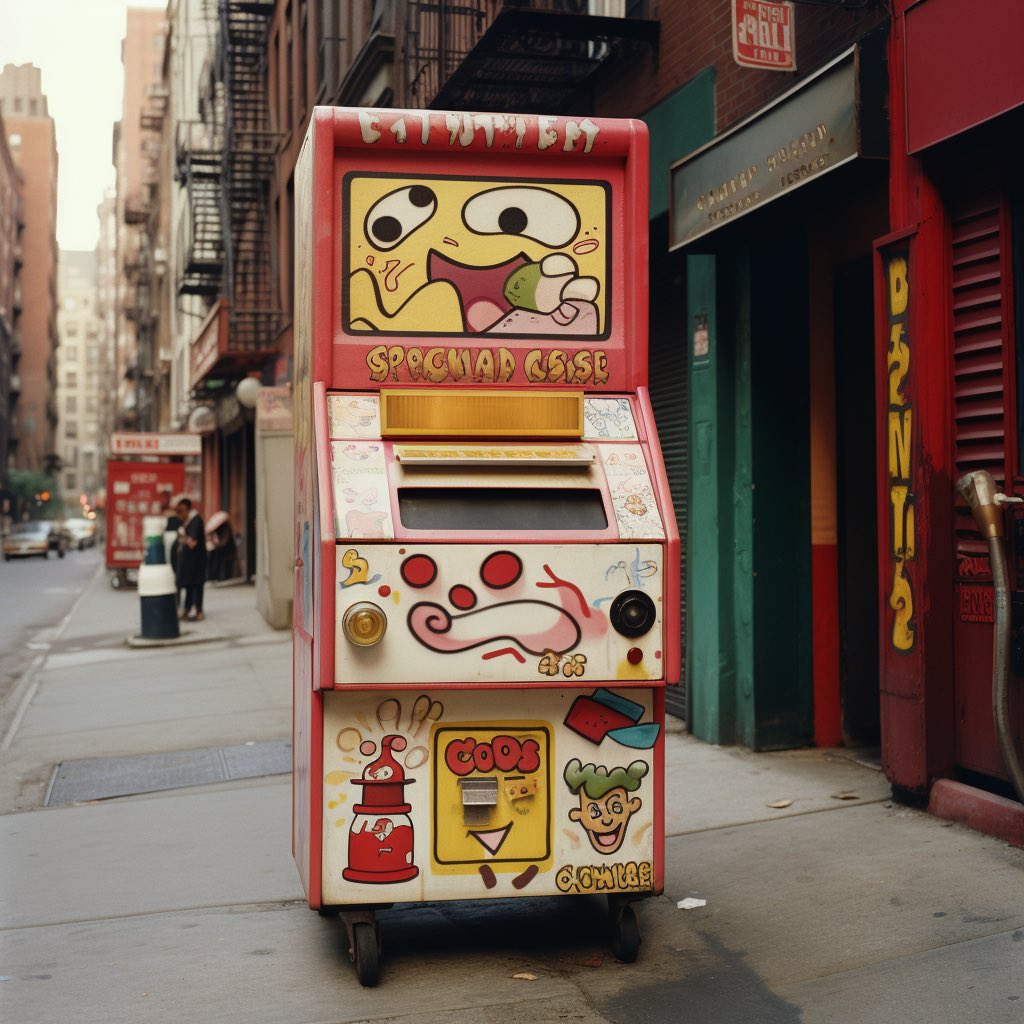}
  		\vspace{0.15cm} \\
	\end{tabular}
	\caption{Some examples of images downloaded from the social network X. From left to right: a real image, synthetic images from DALL·E 3 \cite{dalle3}, Firefly \cite{firefly}, Midjourney \cite{midjourney}.}
 \label{fig:tiwtter_images}
 \end{figure}

\section{Few-shot analysis in the wild}
\label{few-shot}

In this section we present an experiment in a few-shot scenario. The idea is to explore the ability of our detector to work with very limited data and adapt to a situation where only a few real and synthetic examples are available. 
We downloaded 500 real images and 1,500 synthetic images from X from three different generators DALL·E 3, Midjourney and Firefly (some examples can be seen in Fig.~\ref{fig:tiwtter_images}). To understand which generator was used to create a specific image, we relied on tags and annotations present on the social network.

In our few-shot scenario, we take $10$ real images and $10$ generated images as examples from a specific model and test on all the others images (note that we present average results on 1,000 runs). 
Results are reported in Tab.~\ref{tab:Twitter} in terms of AUC and accuracy.
The availability of just 10+10 images of the target data provides an impressive performance boost with respect to the same method trained on a dataset much larger but not aligned with the test data (real images come from COCO, synthetic images from Latent diffusion models).
It is important to underline that this is a realistic scenario, in which one is called to decide on images generated by a new method and a few sample images are available as a support.  
We believe that this can represent an interesting direction for the application in the wild or to easily adapt a detector to more challenging situations where some prior information is available.

To complete our analysis on this dataset we include in Tab.~\ref{tab:Twitter} the comparison with SoTA methods and our original proposal. We can observe that the performance of all the methods degrade which highlights the increased difficulty to handle a realistic scenario over the web. The best performance can be obtained by using our fusion approach that take the best of the low-level and high-level features and is able to achieve on average more than 90\% in terms of AUC which is comparable with the few-shot analysis where some prior knowledge is available.

\begin{table}[t!]
\centering
\resizebox{1.\linewidth}{!}{
\def\arraystretch{0.8}
\begin{tabular}{@{}l|c|c|c|c|c|l@{}} 
\ru Generator                                     & \rob{modality} & \rob{LSUN \cite{yu2015lsun}} & 
\rob{FFHQ \cite{karras2019stylegan}} & \rob{ImageNet \cite{deng2009imagenet}}& \rob{COCO \cite{lin2014microsoft}} & Resolution                 \\ \midrule
              BigGAN    \cite{brock2018large}        & c   &     &     & \cm &     & $256^{2}$, $512^{2}$ \\
                EG3D    \cite{chan2022efficient}     & u   &     & \cm &     &     & $512^{2}$ \\
   Diff.ProjectedGAN    \cite{wang2023diffusion}     & u   & \cm &     &     &     & $256^{2}$ \\
               GALIP    \cite{tao2023galip}          & t   &     &     &     & \cm & $224^{2}$ \\ \midrule
      Taming Transf.    \cite{esser2021taming}       & u,c &     & \cm & \cm &     & $256^{2}$ \\
         DALL·E mini    \cite{dayma2021dallemini}    & t   &     &     &     & \cm & $256^{2}$ \\ \midrule
                DDPM    \cite{ho2020denoising}       & u   &     & \cm &     &     & $256^{2}$ \\
Deepfloyd-IF II stage   \cite{deepfloydif}           & t   &     &     &     & \cm & $256^{2}$ \\
\bottomrule                                                                  
\end{tabular}                                                                
}
\caption{Image generators used in our experiments: GAN-based, VQ-GAN-based and DM-based.
The generation modalities, unconditional (u), conditional (c), and text-to-image (t), is reported in the second column.
The last column reports the resolution of the images in the dataset.} 
\label{tab:sup_datasets}
\end{table}

\renewcommand{\b}[1]{{\bf #1}}
\begin{table*}[t!]
{\small
\centering
\resizebox{1.\linewidth}{!}{
\begin{tabular}{l| C{18mm}C{18mm}C{18mm}C{18mm}C{18mm}C{18mm}C{18mm}C{18mm} |C{18mm} } \toprule
 \ru AUC/Acc  &  \twr{BigGAN}   &  \twr{EG3D} &   Diff. Proj.   &   \twr{GALIP}  &  Taming  &  DALL·E  &  \twr{DDPM}  &  Deepfloyd-IF & \twr{AVG}  \\[-0mm]
 \ru Method   &      &       &    GAN  &     &  Transf.  &  Mini  &   &  II stage    \\
\midrule
\ru Wang et al.  &    92.7  /    66.1  &    94.4  /    59.2  &    89.8  /    52.1  &    89.7  /    57.4  &    54.3  /    51.7  &    62.5  /    51.8  &    31.6  /    50.1  &    43.1  /    50.1  &    69.8  /    54.8  \\
\ru PatchFor.    &    85.5  /    52.5  &    69.8  /    50.0  &    92.6  /    61.7  &    98.2  /    73.4  &    71.2  /    51.0  &    83.8  /    51.4  &    98.4  /    50.2  &    83.4  /    50.0  &    85.4  /    55.0  \\
\ru Grag. et al. &    98.7  /    94.2  &    98.9  /    93.8  & \b{100.} /    72.5  &    96.2  /    79.5  &    90.3  /    76.4  &    83.4  /    62.5  &    49.7  /    45.1  &    71.2  /    61.3  &    86.0  /    73.2  \\
\ru Mand. et al. &    92.2  /    83.0  & \b{100.} / \b{99.8} &    64.5  /    49.9  &    77.6  /    59.1  &    91.8  /    84.1  &    83.6  /    69.8  & \b{99.9} / \b{97.4} &    49.2  /    48.9  &    82.3  /    74.0  \\
\ru Liu et al.   &    94.7  /    81.3  &    99.0  /    86.3  & \b{99.5} /    84.9  &    94.3  /    78.4  &    95.4  /    78.9  &    98.4  /    88.1  &    22.8  /    49.6  &    97.4  /    86.8  &    87.7  /    79.3  \\
\ru Corvi et al. &    83.4  /    51.8  &    25.2  /    50.0  &    96.6  /    71.2  &    87.7  /    50.9  & \b{99.3} /    89.5  & \b{99.7} /    83.8  & \b{100.} /    90.3  &    68.5  /    50.8  &    82.5  /    67.3  \\
\ru LGrad        &    77.2  /    69.0  &    68.8  /    59.8  & \b{99.5} /    90.1  &    56.7  /    55.1  &    74.1  /    64.5  &    67.3  /    59.9  &    ~9.8  /    16.4  &    75.0  /    62.5  &    66.1  /    59.7  \\
\ru Ojha et al.  & \b{99.6} / \b{96.4} &    92.6  /    82.5  &    97.4  /    75.2  &    98.6  /    89.9  &    94.1  /    84.8  &    97.1  /    84.9  &    77.7  /    68.2  &    60.8  /    50.0  &    89.7  /    79.0  \\
\ru DIRE-1       & \b{99.8} /    95.3  &    50.1  /    50.0  &    49.8  /    51.6  & \b{100.} /    96.7  &    73.1  /    72.4  & \b{99.7} /    96.5  &    23.1  /    50.0  & \b{99.4} / \b{95.8} &    74.4  /    76.0  \\
\ru DIRE-2       &    98.6  /    82.4  &    46.1  /    50.0  &    50.2  /    51.4  & \b{99.3} /    81.8  &    77.0  /    66.2  &    98.7  /    81.8  &    14.0  /    49.8  &    95.6  /    80.5  &    72.4  /    68.0  \\
\ru NPR          &    86.8  /    77.2  &    53.3  /    57.5  & \b{100.} / \b{99.2} &    90.7  /    77.6  &    80.2  /    69.2  &    79.0  /    73.3  &    92.4  /    61.6  &    90.9  /    76.8  &    84.2  /    74.1  \\
\midrule
\ru Ours 1k      &    96.9  /    86.1  &    87.0  /    73.0  & \b{99.3} /    70.2  & \b{100.} / \b{99.7} & \b{99.4} /    95.0  & \b{100.} / \b{99.1} &    95.1  /    86.7  & \b{99.7} /    90.7  & \b{97.2} / \b{87.6} \\
\ru Ours 1k+     &    92.0  /    80.0  &    76.4  /    65.5  &    89.6  /    67.7  & \b{99.9} /    98.6  &    93.5  /    83.0  & \b{99.6} /    95.0  &    94.9  /    87.4  & \b{99.7} / \b{96.0} &    93.2  /    84.1  \\
\ru Ours 10k     &    98.2  /    87.0  &    87.7  /    63.2  & \b{99.7} /    87.4  & \b{100.} / \b{99.8} & \b{99.7} / \b{97.4} & \b{100.} / \b{99.4} &    93.9  /    61.9  & \b{99.8} /    85.9  & \b{97.4} /    85.3  \\
\ru Ours 10k+    &    93.1  /    77.5  &    79.9  /    58.8  &    92.5  /    76.0  & \b{100.} /    97.0  &    94.9  /    79.5  & \b{99.9} /    94.4  &    98.1  /    91.5  & \b{99.8} /    92.2  &    94.8  /    83.3  \\
 \bottomrule
\end{tabular}
}
\vspace{-1.5mm}
\caption{{\bf Comparison with SOTA methods on additional data.} The results are in terms of AUC/Accuracy. The entries in bold underline the best performance for each dataset. For our approach we show four variants: trained on 1k real and 1k fake images; 10k real and 10k fake images; trained on 1k and 1k fake images but including compressed/resized images (1k+) and trained on 10k and 10k fake images but including compressed/resized images (10k+).}
\label{tab:other}
}
\end{table*}

\begin{table*}[t!]
{\small
\centering
\resizebox{1.\linewidth}{!}{
\begin{tabular}{l| C{18mm}C{18mm}C{18mm}C{18mm}C{18mm}C{18mm}C{18mm}C{18mm} |C{18mm} } \toprule
 \ru AUC/Acc   &  \twr{BigGAN}   &  \twr{EG3D} &   Diff. Proj.   &  \twr{GALIP}  &  Taming  &  DALL·E  &  \twr{DDPM}  &  Deepfloyd-IF  &  \twr{AVG} \\[-0mm]
 \ru Method      &       &   &    GAN   &    &  Transf.  &  Mini  &   &  II stage    \\
  \midrule
\ru Wang et al.    &    82.5  /    55.5  &    84.7  /    52.2  &    80.8  /    53.2  &    92.2  /    59.2  &    66.2  /    50.6  &    66.7  /    50.5  &    69.6  /    50.2  &    47.9  /    49.9  &    73.8  /    52.7  \\
\ru PatchFor.      &    58.7  /    50.3  &    60.3  /    50.0  &    55.6  /    50.0  &    72.3  /    50.3  &    50.4  /    50.1  &    71.0  /    50.1  &    82.9  /    51.0  &    69.7  /    50.0  &    65.1  /    50.2  \\
\ru Grag. et al.   & \b{97.6} /    74.4  & \b{92.7} /    56.5  & \b{99.2} /    65.6  & \b{99.1} /    79.8  &    83.6  /    53.3  &    89.6  /    54.0  &    74.5  /    49.4  &    54.3  /    50.0  &    86.3  /    60.4  \\
\ru Mand. et al.   &    71.4  /    55.5  &    85.8  / \b{62.1} &    60.3  /    49.6  &    78.1  /    57.3  &    82.3  /    59.7  &    68.2  /    57.4  &    61.1  /    58.6  &    49.0  /    49.8  &    69.5  /    56.3  \\
\ru Liu et al.     &    57.7  /    52.0  &    51.6  /    50.4  &    62.9  /    54.1  &    65.2  /    52.3  &    49.2  /    50.2  &    49.1  /    50.7  &    58.3  /    49.1  &    64.4  /    52.0  &    57.3  /    51.4  \\
\ru Corvi et al.   &    77.3  /    53.0  &    64.5  /    59.7  &    93.3  /    59.8  &    87.4  /    51.5  & \b{94.4} /    76.7  &    97.4  /    74.8  &    74.7  /    59.6  &    70.4  /    50.9  &    82.4  /    60.7  \\
\ru LGrad          &    53.5  /    51.6  &    53.2  /    51.1  &    56.9  /    51.1  &    57.0  /    50.3  &    51.3  /    51.1  &    41.7  /    49.1  &    56.0  /    49.7  &    38.7  /    48.8  &    51.0  /    50.4  \\
\ru Ojha et al.    &    94.8  / \b{84.9} &    79.3  / \b{61.5} &    84.7  / \b{68.5} &    95.7  /    84.5  &    92.9  / \b{81.8} &    89.1  /    72.8  &    90.8  /    79.1  &    59.1  /    49.5  &    85.8  /    72.8  \\
\ru DIRE-1         &    47.2  /    49.5  &    35.2  /    49.3  &    35.2  /    49.6  &    47.4  /    50.6  &    47.3  /    49.7  &    54.6  /    51.1  &    38.5  /    49.4  &    64.6  /    53.2  &    46.2  /    50.3  \\
\ru DIRE-2         &    44.6  /    47.0  &    33.6  /    47.1  &    37.5  /    44.9  &    46.2  /    47.2  &    43.3  /    46.8  &    53.1  /    50.0  &    42.2  /    47.5  &    66.3  /    59.3  &    45.9  /    48.7  \\
\ru NPR            &    52.6  /    50.0  &    40.2  /    49.8  &    48.8  /    50.3  &    57.8  /    50.6  &    46.2  /    49.8  &    51.7  /    49.6  &    54.5  /    50.5  &    51.3  /    48.9  &    50.4  /    49.9  \\
\midrule
\ru Ours 1k        &    84.6  /    75.8  &    64.8  /    60.3  &    73.4  /    66.7  &    98.3  /    92.6  &    85.1  /    75.8  &    96.2  /    88.9  &    76.1  /    67.7  &    90.0  /    79.9  &    83.6  /    76.0  \\
\ru Ours 1k+       &    89.6  /    79.7  &    60.5  /    55.1  &    69.6  /    61.0  & \b{99.9} / \b{98.4} &    86.9  /    78.0  & \b{98.9} /    93.7  & \b{96.4} /    84.8  & \b{99.5} / \b{96.5} &    87.7  /    80.9  \\
\ru Ours 10k       &    84.1  /    67.0  &    64.3  /    54.9  &    73.7  /    62.9  &    98.1  /    83.9  &    86.2  /    74.0  &    96.8  /    77.8  &    70.9  /    60.7  &    91.3  /    67.5  &    83.1  /    68.6  \\
\ru Ours 10k+      &    92.4  /    82.4  &    63.1  /    55.5  &    74.1  /    63.9  & \b{99.9} / \b{98.9} &    88.1  /    78.2  & \b{99.4} / \b{94.9} & \b{96.1} / \b{87.1} & \b{99.7} / \b{97.0} & \b{89.1} / \b{82.2} \\
 \bottomrule
\end{tabular}
}
\vspace{-1.5mm}
\caption{{\bf Comparison with SOTA methods on additional data in the presence of post-processing.} The results are in terms of AUC/Accuracy. The entries in bold underline the best performance for each dataset.}
\label{tab:other_postprocessing}
}
\end{table*}

\section{Further generalization results}
\label{generalization}

In this Section we extend our generalization analysis to additional data
and consider 8 more generators with 8,000 additional synthetic images for our test set (Tab.~\ref{tab:sup_datasets}).
In Tab.~\ref{tab:other} we show the results in terms of AUC and accuracy on such data  for the four version of the proposed CLIP-based detector and for the SoTA methods described in Section \ref{sota_methods}.
We can observe that SoTA methods present a larger variability in terms of performance, with very good results on some generators and very bad on others. Instead, our method can provide consistently good performance over all the data both in terms of AUC and accuracy, with an average gain over the best reference of around 8.5$\%$ in terms of AUC and 13$\%$ in terms of Accuracy.
This difference is even more noticeable on data that have undergone random compression and resizing (Tab.~\ref{tab:other_postprocessing}). In fact, for most of the competitors there is a dramatic performance loss, sometimes to a random guess level. Some other methods, such as, Ohja et al., preserve a good performance and the same happens for ours. Of course, in this more challenging scenario our best variant is the one that includes some form of augmentation in the reference data but even the variant trained on the original data provides satisfying results.

\end{appendix}

{
    \small
    \bibliographystyle{ieeenat_fullname}
    \bibliography{main}
}

\end{document}